\documentclass{article}

\usepackage{microtype}
\usepackage{graphicx}
\usepackage{subfigure}
\usepackage{overpic}
\usepackage{booktabs} 

\usepackage{hyperref}

\usepackage[accepted]{icml2024}

\usepackage{amsmath}
\usepackage{amssymb}
\usepackage{mathtools}
\usepackage{amsthm}
\usepackage{bm}
\usepackage{multirow}

\usepackage[capitalize,noabbrev]{cleveref}

\theoremstyle{plain}

\theoremstyle{definition}

\theoremstyle{remark}

\usepackage[textsize=tiny]{todonotes}

\makeatletter
\newcommand{\@giventhatstar}[2]{$\left(#1\,\middle|\,#2\right)$}
\newcommand{\@giventhatnostar}[3][]{#1(#2\,#1|\,#3#1)}
\newcommand{\giventhat}{\@ifstar\@giventhatstar\@giventhatnostar}
\makeatother

\newcommand{\dist}[3]{#1\giventhat{#2}{#3}}
\newcommand{\code}[1]{\texttt{#1}}

\newcommand{\expnumber}[2]{{#1}\mathrm{e}{#2}}
\DeclarePairedDelimiterX{\infdivx}[2]{[}{]}{%
  #1\,\delimsize\|\,#2%
}
\newcommand{\kldiv}{\text{KL}\infdivx}

\makeatletter
\providecommand\citet[2][]{%
  \edef\@tempa{#1}
  \citeauthor{#2} (%
    \citeyear{#2}%
    \ifx\@empty\@tempa\else,~#1\fi
  )
}
\makeatother

\icmltitlerunning{Sequential Disentanglement by Extracting Static Information From A Single Sequence Element}

\begin{document}

\twocolumn[
\icmltitle{Sequential Disentanglement by Extracting Static Information From A Single Sequence Element}



\icmlsetsymbol{equal}{*}

\begin{icmlauthorlist}
\icmlauthor{Nimrod Berman}{equal,yyy}
\icmlauthor{Ilan Naiman}{equal,yyy}
\icmlauthor{Idan Arbiv}{equal,yyy}
\icmlauthor{Gal Fadlon}{equal,yyy}
\icmlauthor{Omri Azencot}{yyy}
\end{icmlauthorlist}

\icmlaffiliation{yyy}{Department of Computer Science, Ben-Gurion University of the Negev, Beer-Sheva, Israel}

\icmlcorrespondingauthor{Nimrod Berman}{bermann@post.bgu.ac.il}

\icmlkeywords{Machine Learning, ICML}

\vskip 0.3in
]



\printAffiliationsAndNotice{\icmlEqualContribution} 

\begin{abstract}

One of the fundamental representation learning tasks is unsupervised sequential disentanglement, where latent codes of inputs are decomposed to a single static factor and a sequence of dynamic factors. 
To extract this latent information, existing methods condition the static and dynamic codes on the entire input sequence. 
Unfortunately, these models often suffer from \emph{information leakage}, i.e., the dynamic vectors encode both static and dynamic information, or vice versa, leading to a non-disentangled representation. Attempts to alleviate this problem via reducing the dynamic dimension and auxiliary loss terms gain only partial success. Instead, we propose a novel and simple architecture that mitigates information leakage by offering a simple and effective subtraction inductive bias while conditioning on a \emph{single} sample. Remarkably, the resulting variational framework is simpler in terms of required loss terms, hyper-parameters, and data augmentation. We evaluate our method on multiple data-modality benchmarks including general time series, video, and audio, and we show beyond state-of-the-art results on generation and prediction tasks in comparison to several strong baselines. Code is at \href{https://github.com/azencot-group/DBSE}{GitHub}.


\end{abstract}

\section{Introduction}

Modern representation learning~\citep{bengio2013representation, scholkopf2021toward} identifies unsupervised disentanglement as one of its fundamental challenges, where the main goal is to decompose input data to its latent factors of variation. Separating the learned representation into independent factors can improve multiple machine learning tasks from three aspects: (1) explainability, (2) generalizability (3) controllability. Capturing distinct variation allows filtering out undesired variations, reducing the sample complexity of downstream learning~\citep{villegas2017decomposing, denton2017unsupervised}, and facilitating more controllable generations \citep{tian2021a}. Further, disentangled representations are instrumental in numerous applications including classification~\citep{locatello2020disentangling}, prediction~\citep{hsieh2018learning}, and interpretability~\citep{higgins2016beta, naiman2023operator}, to name just a few. In the sequential case, inputs are typically split to a single static (time-invariant) factor encoding features that do not change over time, and to multiple dynamic (time-varying) components, one per sample. For instance, in a smiling face video, the latent representation can be disentangled into a static component encoding the person's identity (time-invariant factor) and a dynamic component encoding the smiling motion (time-variant factor). Such disentangled representations hold promise for various downstream tasks such as classification, retrieval, and synthetic video generation with style transfer.

Existing sequential disentanglement works are commonly based on variational autoencoders (VAEs)~\cite{kingma14auto} and their dynamic extensions~\citep{girin2021dynamical}. To model the variational posterior, \citet{li2018disentangled, bai2021contrastively} and others condition the static and dynamic factors on the entire input sequence. However, under this modeling perspective, highly expressive deep encoder modules struggle with \emph{information leakage} problems. Namely, the learned time-varying components capture dynamic as well as static information, whereas the static vector encodes non-meaningful features, or vice versa. To resolve this issue, \citet{li2018disentangled} propose to reduce the dynamic dimension drastically, and \citet{zhu2020s3vae,bai2021contrastively} introduce additional mutual information (MI) loss terms. While these approaches generally improved disentanglement and leakage issues, new challenges emerged. First, small time-varying vectors are limited in modeling complex dynamics. Second, sampling good positive and negative examples to estimate mutual information is hard \citep{naiman2023sample}. Finally, optimizing models with multiple losses and balancing their MI penalties is difficult. These challenges raise the question: can we design an unsupervised sequential disentanglement framework that avoids the above complexities and avoids information leakage?

To induce deep neural networks with a certain behavior, prior successful attempts opted for designing tailored architectures, often arising from an overarching assumption. For instance, convolutional neural networks~\citep{lecun1989backpropagation} use shared kernel filters across natural images, exploiting their translation-invariance structure. Similarly, natural language processing methods~\citep{bahdanau2014neural} employ attention modules, assuming not all words in the source sentence share the same effect on the target sentence. Inspired by these successes, we suggest a novel sequential disentanglement model that drastically alleviates information leakage, based on the assumption that the static posterior can be conditioned on a \emph{single} sample from the sequence. Further, our modeling guides architecture design, where we eliminate static features from the dynamic factors, and static and dynamic codes are learned separately from one another. The resulting framework has no MI terms and thus less hyper-parameters, and further, it requires no constraints on the dimension of factors. 

Our main contributions can be summarized as follows:
\begin{enumerate}
    \item We introduce a novel sequential disentanglement model whose static posterior is conditioned on a single series element, leading to a new neural architecture that learns the static code independently from the sequence, and subtracts its contents from the dynamic factors.

    \item Our method does not restrict the dimension of disentangled factors, and thus, it supports complex dynamics. Further, it mitigates information leakage without mutual information loss terms, yielding a simple training objective with only two hyper-parameters.

    \item We extensively evaluate our approach both qualitatively and quantitatively on several data modalities obtaining state-of-the-art results on standard challenging benchmarks. Additionally, we extend current benchmarks to assess information leakage issues, and we show the superiority of our framework in mitigating information leakage in comparison to SOTA techniques.
\end{enumerate}

\section{Related Work}


\paragraph{Unsupervised disentanglement.} A large body of work is dedicated to studying unsupervised disentanglement using VAEs~\citep{kingma14auto}. For instance, \citet{higgins2016beta} augment VAEs with a hyper-parameter weight on the Kullback--Liebler divergence term, leading to improved disentanglement. In~\cite{kumar2017variational}, the authors regularize the expectation of the posterior, and \citet{bouchacourt2018multi} learn multi-level VAE by grouping observations. \citet{kim2018disentangling} promote independence across latent dimensions, and \citet{chen2018isolating} decompose the objective function and identify a total correlation term. In addition, generative adversarial networks (GANs) are used to maximize mutual information~\cite{chen2016infogan} and to compute an attribute dependency metric~\cite{wu2021stylespace}.

\paragraph{Sequential disentanglement.} Early probabilistic models suggest conditioning on the mean of past features~\citep{hsu2017unsupervised}, and directly on the features~\cite{li2018disentangled}. Disentangling video sequences is achieved using generative adversarial networks~\citep{villegas2017decomposing, tulyakov2018mocogan} and a recurrent model with an adversarial loss~\citep{denton2017unsupervised}. To address information leakage issues, \citet{zhu2020s3vae} introduce modality-based auxiliary tasks and supervisory signals, whereas \citet{bai2021contrastively} utilize contrastive estimation using positive and negative examples. Optimal transport and the Wasserstein distance are used to regularize the evidence lower bound in~\citep{han2021disentangled}. \citet{tonekaboni2022decoupling} extract local and global representations to encode non-stationary time series data. Recently, \citet{naiman2023sample} developed a modality-independent method to sample latent positive and negative examples. Others considered the more general multifactor disentanglement, where every sequence is decomposed to multiple static and dynamic factors~\citep{bhagat2020disentangling, yamada2020disentangled, berman2023multifactor}.

\section{Background}
\label{sec:background}

\paragraph{Problem formulation.} We generally follow the notation and terminology introduced in~\citep{li2018disentangled}. Let $x_{1:T} = \{ x_1, \dots, x_T\}$ denote a multivariate sequence of length $T$, where $x_t \in \mathbb{R}^d$ for every $t$. Given a dataset $\mathcal{D} = \{ x_{1:T}^j \}_{j=1}^N$, the goal of sequential disentanglement is to extract an alternative representation of $x_{1:T}$, where we omit $j$ for brevity, via a static (time-invariant) factor $s$ and multiple dynamic (time-varying) components $d_{1:T}$. For instance, given a sequence of frames depicting a person making a facial expression, we expect $s$ to encode the identity of the person, and $d_{1:T}$ to capture the expressions as they change along the sequence. Importantly, we note that $s$ is shared across $x_{1:T}$ as the person's identity remains fixed. 

\begin{figure*}[t]
  \centering
  \begin{overpic}[width=1\linewidth]{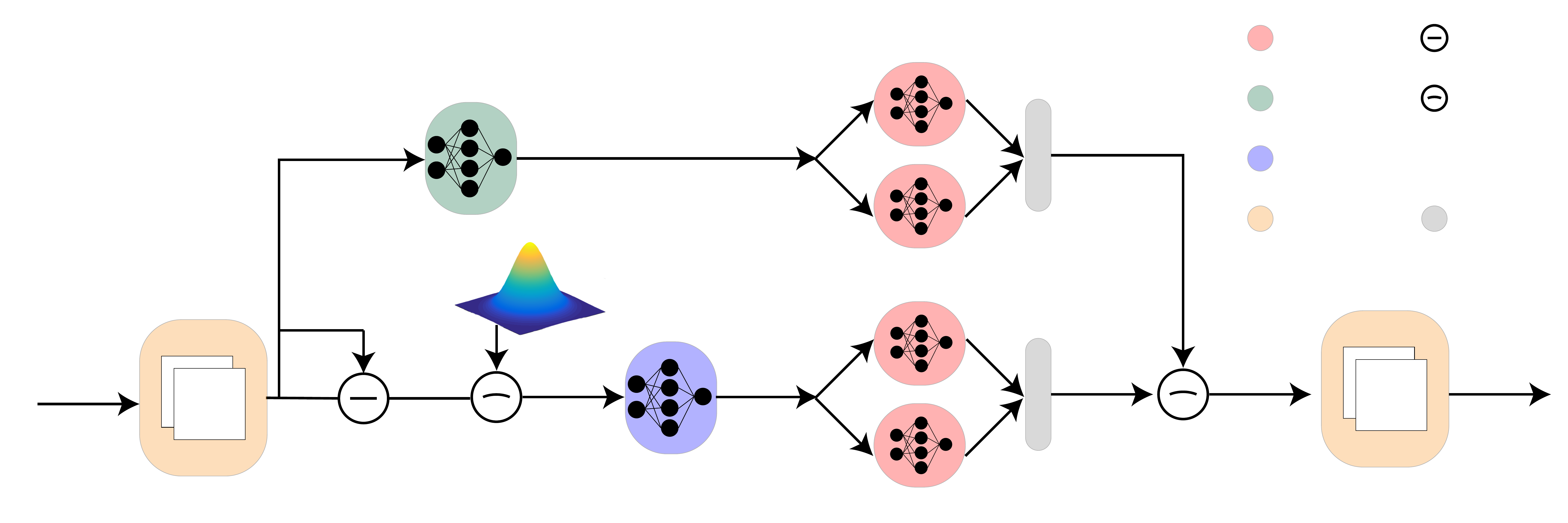}

        \put(3.5, 6){\small $x_{1:T}$}
        \put(18.5, 13){\small $g_1$} \put(18.35, 6.5){\small $g_{2:T}$}
        \put(26.75, 6.4){\small $\tilde{g}_{2:T}$}
        \put(29.2, 11.25){\small $\tilde{g}_1$}
        \put(40, 21.5){\small $\tilde{s}$}
        \put(36, 6.25){\small $\tilde{g}_{1:T}$}
        \put(46.5, 6.25){\small $h_{1:T}$}
        \put(70.5, 21.95){\small $s$} 
        \put(68, 6.5){\small $d_{1:T}$}
        
        \put(81.6, 30.5){\small linear} \put(81.6, 26.5){\small MLP} \put(81.6, 22.6){\small LSTM} \put(81.6, 19){\small enc/dec} \put(92.75, 19){\small repr. trick}
        \put(92.75, 30.5){\small subtract}
        \put(92.75, 26.6){\small concat}
  \end{overpic}
  \vspace{-5mm}
  \caption{Our network is composed of an encoder (left), a decoder (right) and two paths in-between for computing the static factor (top) and the dynamic components (bottom). For full architecture details, see App.~\ref{app:architecture}.}
  \label{fig:arch}
\end{figure*}
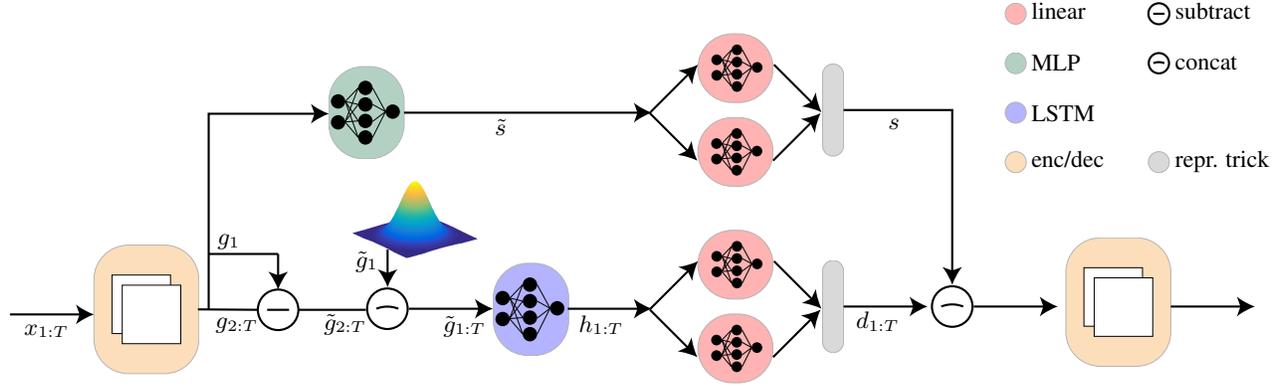

\paragraph{Sequential probabilistic modeling.} Typically, the static and dynamic features are assumed to be independent, and thus the joint distribution $P := p(x_{1:T}, z \, ; \, \theta, \psi)$ is given by
\begin{equation} \label{eq:vae_prior}
    P = \left[p(s) \prod_{t=1}^T \dist{p}{d_t}{d_{<t} \, ; \, \psi} \right] \cdot \prod_{t=1}^T \dist{p}{x_t}{s, d_t \, ; \, \theta} \ ,
\end{equation}
where $z := (s,d_{1:T})$ combines static and dynamic components, $d_t$ depends on prior features $d_{<t}$, and $x_t$ can be reconstructed from the static and current dynamic codes. The static prior distribution is modeled by a standard Gaussian distribution $p(s) := \mathcal{N}(0, I)$, whereas the dynamic prior is computed via a recurrent neural network $\dist{p}{d_t}{d_{<t} \, ; \, \psi} := \mathcal{N}(\mu(d_{<t}), \sigma^2(d_{<t}) \, ; \, \psi)$ to capture nonlinear dynamics~\citep{chung2015recurrent, naiman2024generative}. Exploiting the independence between time-varying and time-invariant factors, the posterior $Q := \dist{q}{s, d_{1:T}}{x_{1:T} \, ; \, \phi}$ is parameterized by $\phi=(\phi_s, \phi_d)$, and it reads
\begin{equation} \label{eq:vae_app_posterior}
    Q = \dist{q}{s}{x_{1:T} \, ; \, \phi_s} \prod_{t=1}^T \dist{q}{d_t}{d_{<t}, x_{\leq t} \, ; \, \phi_d} \ .
\end{equation}

\paragraph{Information preference and information leakage.} \emph{Information preference} \cite{chen2016variational, zhao2019infovae}, is a property where the input and the latent codes lack of mutual information. In the context of sequential disentanglement, information preference means that the MI between latent codes and the input is low. A separate problem is \emph{information leakage}, where static information is leaked into the dynamic code or vice versa. Formally, $I_q(s,X_d) > 0$ or $I_q(d, X_s) > 0$. Where $X_d$ and $X_s$ respectively defined as the ground truth static and dynamic information of the data, and $I_q(\cdot, \cdot)$ denotes mutual information.

\paragraph{Objective function.} The corresponding evidence lower bound related to Eqs.~\ref{eq:vae_prior} and \ref{eq:vae_app_posterior} is given by 
\begin{equation}  \label{eq:dsvae_elbo}
\max_{\theta,\phi,\psi} \; \mathbb{E}_{p_\mathcal{D}} \left[\mathbb{E}_{z \sim q_\phi} L - \beta K \right] ,
\end{equation}
where $p_\mathcal{D}$ is the train set distribution, $q_\phi := \dist{q}{z}{x_{1:T} \, ; \, \phi}$, $\beta \in \mathbb{R}^+$ is a balancing scalar, $L := \log \dist{p}{x_{1:T}}{z \, ; \, \theta}$, and $K := \kldiv{\dist{q}{z}{x_{1:T} \, ; \, \phi}}{p(z \, ; \, \psi)}$.

Unfortunately, the above model is prone to information leakage and information preference. To alleviate these issues, \citet{li2018disentangled} decreased the dimension of $d_t$ such that $\text{dim}(d_t) \ll \text{dim}(s)$. The intuition behind this heuristic is to provide the model with a minimal subspace for learning dynamics, without extra degrees of freedom to capture the static information. Other approaches~\citep{zhu2020s3vae, bai2021contrastively} aim to mitigate these problems by augmenting the objective function with mutual information terms, minimizing $I_q(s; d_{1:T})$ and maximizing $I_q(s; x_{1:T})$ and $I_q(d_{1:T}; x_{1:T})$. However, evaluating $I_q(u; v)$ is difficult \citep{chen2018isolating}, and it is approximated by contrastive estimation~\citep{oord2018representation} which requires positive and negative data samples. Overall, while the above solutions lessen the detrimental effects of the mentioned problems, several challenges still exist. 1) the hyper-parameter $\text{dim}(d_t)$ may be difficult to tune, and it limits model expressivity in capturing complex dynamics; 2) contrastive estimation is often implemented with domain-dependent data augmentation, hindering its use on arbitrary data modalities~\citep{tonekaboni2022decoupling}; and 3) models with multiple loss terms (including MI) may be sensitive to hyper-parameter choices and their optimization is computationally expensive.


\section{Method}

Motivated by the challenges described in Sec.~\ref{sec:background}, our main goal is to answer the question: 
\begin{center} \vspace{-2mm}
``\textit{Can we perform sequential disentanglement by a biased architecture and a simpler loss?}''
\end{center} \vspace{-2mm}
To this end, we make the following two observations: (i) the approximate posterior distribution is ultimately responsible for extracting latent factors of variation from input sequences, and thus, information leakage issues mainly appear in the posterior; and (ii) the static features are, by definition, time-invariant and shared across the sequence, and therefore, they could be extracted from a single sample in the sequence. Based on these simple observations, we propose a new posterior distribution that conditions the static factor on one sequential element, e.g., on some item $x_i$ where $i \in \{1, \dots, T\}$. Intuitively, $x_i$ may be viewed as an \emph{anchor} example, from which we extract the static information. Moreover, we assume that the time-invariant factor is not required in modeling the time-varying components. Formally, our posterior distribution is defined as follows
\begin{equation} \label{eq:our_app_posterior}
\begin{split}
    &\dist{q}{s, d_{1:T}}{x_{1:T} \, ; \, \phi} = q_{\phi_s} \cdot q_{\phi_d} := \\
    &\dist{q}{s}{x_i \, ; \, \phi_s} \cdot \prod_{t=1}^T \dist{q}{d_t}{d_{<t}, x_{\leq t} \, ; \, \phi_d} \ ,
\end{split}
\end{equation}
where we set $\phi = (\phi_s, \phi_d)$, that is, the static and dynamic codes are computed using neural networks parameterized by $\phi_s$ and $\phi_d$, respectively.

\begin{table*}[!t]
    \caption{Benchmark disentanglement metrics on Sprites and MUG. Results with standard deviation appear in Tab.~\ref{tab:gen_sprites_mug_std}. Arrows denote whether higher or lower results are better.}
    \vskip 0.1in
    \label{tab:gen_sprites_mug}
    \centering
    \footnotesize
        \begin{tabular}[t]{lcccc|cccc}
            \toprule
            & \multicolumn{4}{c|}{Sprites} & \multicolumn{4}{c}{MUG} \\
            Method & Acc$\uparrow$ & IS$\uparrow$ & $H(y|x){\downarrow}$ & $H(y){\uparrow}$ & Acc$\uparrow$ & IS$\uparrow$ & $H(y|x){\downarrow}$ & $H(y){\uparrow}$ \\
            \midrule
            MoCoGAN & $92.89\%$ & $8.461$ & $0.090$ & $2.192$ & $63.12\%$ & $4.332$ & $0.183$ & $1.721$ \\
            DSVAE & $90.73\%$ & $8.384$ & $0.072$ & $2.192$ & $54.29\%$ & $3.608$ & $0.374$ & $1.657$ \\
            R-WAE & $98.98\%$ & $8.516$ & $0.055$ & $\boldsymbol{2.197}$ & $71.25\%$ & $5.149$ & $0.131$ & $1.771$ \\
            S3VAE & $99.49\%$ & $8.637$ & $0.041$ & $\boldsymbol{2.197}$ & $70.51\%$ & $5.136$ & $0.135$ & $1.760$ \\
            SKD & $\boldsymbol{100\%}$ & $\boldsymbol{8.999}$ & $\boldsymbol{\expnumber{1.6}{-7}}$ & $\boldsymbol{2.197}$ & $77.45\%$ & $5.569$ & $0.052$ & $1.769$ \\
            C-DSVAE & $99.99\%$ & $8.871$ & $0.014$ & $\boldsymbol{2.197}$ & $81.16\%$ & $5.341$ & $0.092$ & $1.775$ \\
            SPYL & $100\%$ & $8.942$ & $0.006$ & $\boldsymbol{2.197}$ & $85.71\%$ & $5.548$ & $0.066$ & $1.779$ \\
            \midrule
            Ours & $\boldsymbol{100\%}$ & $8.942$ & $0.006$ & $\boldsymbol{2.197}$ & $\boldsymbol{86.90\%}$ & $\boldsymbol{5.598}$ & $\boldsymbol{0.041}$ & $\boldsymbol{1.782}$ \\
            \bottomrule
        \end{tabular}
        \vskip -0.1in
\end{table*}

In addition to the posterior above, our model uses the same prior distribution as in Eq.~\ref{eq:vae_prior}. Further, given the unique role of $x_i$ in our posterior, we also need to describe the modifications to the loss function in Eq.~\ref{eq:dsvae_elbo}. Specifically, the reconstruction term is split to
\begin{align} \label{eq:our_recon_loss}
\begin{split}
    \mathcal{L}_\text{recon} &= \mathbb{E}_{s \sim q_{\phi_s}} [ \mathbb{E}_{d_{2:T} \sim q_{\phi_d}} \log \dist{p}{x_{2:T}}{s, d_{2:T} \, ; \, \theta}] \\
    &+ \alpha \, \mathbb{E}_{s \sim q_{\phi_s}} [\log \dist{p}{x_1}{s, d_1 \, ; \, \theta} ] \ ,
\end{split}
\end{align}
where, w.l.o.g and to simplify notations we choose $i=1$, i.e., $x_i := x_1$. Notably, $\alpha \in \mathbb{R}^+$ could potentially be taken as $1$. However, we often got better results when $\alpha \neq 1$. The regularization KL term can be elaborated as follows
\begin{align} \label{eq:our_reg_loss}
\begin{split}
    \mathcal{L}_\text{reg} &= \beta \; \kldiv{\dist{q}{s}{x_1 \, ; \, \phi_s}}{p(s)} \\
    &+ \beta \; \kldiv{\dist{q}{d_{1:T}}{x_{1:T} \, ; \, \phi_d}}{p(d_{1:T} \, ; \, \psi)} \ ,
\end{split}
\end{align}
with $\beta \in \mathbb{R}^+$, $\dist{q}{d_{1:T}}{x_{1:T} \, ; \, \phi_d} = q_{\phi_d}$ and $p(d_{1:T} \, ; \, \psi) = \prod_{t=1}^T \dist{p}{d_t}{d_{<t} \, ; \, \psi}$ via Eq.~\ref{eq:vae_prior}. Overall, combining Eqs.~\ref{eq:our_recon_loss} and \ref{eq:our_reg_loss} yields the following total objective function
\begin{equation}  \label{eq:our_elbo}
\mathcal{L} = \max_{\theta,\phi,\psi} \; \mathbb E_{p_D} (\mathcal{L}_\text{recon} - \mathcal{L}_\text{reg}) \ .
\end{equation}

\paragraph{An architectural bias.} Our sequential disentanglement model takes the sequence $x_{1:T}$ as input, and it outputs its reconstruction. The inputs are processed by a problem-dependent encoder, yielding an intermediate sequential representation $g_{1:T}$. Notice, that this module is not recurrent, and thus every $x_t$ is processed independently of other $x_u$ where $u \neq t$. The sequence $g_{1:T}$ is split into two parts $g_1$ (or any other $g_i$) and $g_{2:T}$, undergoing two paths. To extract the static information, $g_1$ is passed to a simple multilayer perceptron (MLP) consisting of a linear layer and \code{tanh} activation, producing $\tilde{s}$. Then, two linear layers learn the mean $\mu(\tilde{s})$ and variance $\sigma^2(\tilde{s})$, allowing to sample the static factor $s$ via the reparametrization trick. We define $\tilde{g}_1 \sim \mathcal{N}(0, I)$. To extract the dynamic information, $\{\tilde{g}_1,g_{2:T} - g_1\}$ is fed to a long short-term memory (LSTM) module~\citep{hochreiter1997long} with a hidden state $h_t$. We can also consider Lipschitz recurrent networks~\citep{erichson2021lipschitz}. Subtracting $g_1$ from $g_{2:T}$ in the same latent manifold mitigates information leakage by ``removing'' static features that exist in $g_{2:T}$, and thus limiting the ability of the LSTM module to extract time-invariant information.  We use $h_{1:T}$ to compute $\mu(h_t)$ and $\sigma^2(h_t)$ using two linear layers, and we sample $d_{1:T}$ by the reparametrization trick. The static and dynamic factors are combined $(s, d_t)$ and passed to a domain-dependent decoder, to produce the reconstruction of $x_t$. See Fig.~\ref{fig:arch} for an annotated scheme of our model.

\section{Experiments}


\subsection{Experimental Setup}

We perform an extensive qualitative, quantitative, and ablation evaluation of our method on datasets of different modalities: video sequences, general time series, and audio recordings. For video sequences, we use the \textbf{Sprites} dataset that contains moving cartoon characters~\citep{reed2015deep}, and the \textbf{MUG} dataset that includes several subjects with various facial expressions~\citep{aifanti2010mug}. The general time series datasets are \textbf{PhysioNet}, consisting of medical records of patients~\citep{goldberger2000physiobank}, \textbf{Air Quality} which includes measurements of air pollution~\citep{zhang2017cautionary}, and \textbf{ETTh1} that measures the electricity transformer temperature~\citep{zhou2021informer}. For audio recordings, we consider the \textbf{TIMIT} dataset, consisting of read speech of short sentences~\citep{garofolo1993timit}. The evaluation tests and tasks we consider serve as the standard benchmark for sequential disentanglement. We compare our approach to recent state-of-the-art (SOTA) techniques including FHVAE \citep{hsu2017unsupervised}, DSVAE \citep{li2018disentangled}, MoCoGan \citep{tulyakov2018mocogan}, R-WAE \citep{han2021disentangled}, S3VAE \citep{zhu2020s3vae}, C-DSVAE \citep{bai2021contrastively}, SKD \citep{berman2023multifactor}, GLR \citep{tonekaboni2022decoupling}, and SPYL \citep{naiman2023sample}. See the appendices for more details regarding datasets (App.~\ref{app:datasets}), evaluation metrics (App.~\ref{app:metrics}), training hyper-parameters (App.~\ref{app:hyperparameters}), and neural architectures (App.~\ref{app:architecture}).

\subsection{Quantitative Evaluation}
\label{subsec:quant_eval}

\begin{table*}
    \caption{Comparing information leakage of static and dynamic features in MUG dataset.}
    \vskip 0.1in
    \label{tab:full_dis_met_cls_task}
    \resizebox{1\textwidth}{!}{
    \centering
    \begin{tabular}[t]{ll|ccc|ccc}
        \toprule
        & & \multicolumn{3}{c|}{Static Features} & \multicolumn{3}{c}{Dynamic Features} \\
        Classifier & Method  & Static Acc $\uparrow$ & Dynamic Acc $\downarrow$ & Leakage Gap $\uparrow$ & Static Acc  $\downarrow$ & Dynamic Acc $\uparrow$ & Leakage Gap $\uparrow$ \\
        \midrule
        \multirow{7}{*}{Generation} & random & - & $16.66\%$ & - & $1.92 \%$ & - & - \\
        & C-DSVAE    & $99.12\% $ & $29.9\% $ & $69.22\%$ & $3.75\%$ & $81.16\%$ & $77.41\%$ \\
        & SPYL       & $\boldsymbol{99.45\%} $ & $27.65\% $ & $71.8\%$ & $3.63\%$ & $85.71\%$ & $82.08\%$ \\
        & \textbf{Ours}        & $99.42\% $ & $\boldsymbol{20.85\%}$ & $\boldsymbol{\textcolor{blue}{78.57\%}}$ & $\boldsymbol{2.89\%}$ & $\boldsymbol{86.90\%}$ & $\boldsymbol{\textcolor{blue}{84.01\%}}$ \\
        \cmidrule{2-8}
        & Ours w.o. loss    & $52.97\% $ & $18.93\% $ & $34.04\%$ & $2.58\%$ & $66.28\%$ & $63.70\%$ \\
        & Ours w.o. sub    & $34.41\% $ & $19.22\% $ & $15.19\%$ & $39.82\%$ & $83.31\%$ & $43.18\%$ \\
        & Ours w.o. both    & $10.94\% $ & $17.86\% $ & $6.92\%$ & $15.76\%$ & $71.95\%$ & $56.19\%$ \\
                \midrule[2pt]
        \multirow{4}{*}{Latent} & random & - & $16.66\%$ & - & $1.92 \%$ & - & - \\
        & C-DSVAE    & $98.75\%$ & $76.25\% $ & $22.25\%$ & $26.25\%$ & $82.50\%$ & $56.25\%$ \\
        & SPYL       & $98.12\% $ & $68.75\%$ & $29.37\%$ & $\boldsymbol{10.00\%}$ & $\boldsymbol{85.62\%}$ & $\boldsymbol{\textcolor{blue}{75.62\%}}$ \\
        & Ours        & $\boldsymbol{99.35\%}$ & $\boldsymbol{45.06\%}$ & $\boldsymbol{\textcolor{blue}{54.29\%}}$ & $11.36\%$ & $85.51\%$ & $74.15\%$ \\
        \bottomrule
    \end{tabular}
    }
    \vskip -0.1in
\end{table*}

\subsubsection{Video And Audio Sequences}

\paragraph{Disentangled generation.} We follow sequential disentanglement works and their protocol to quantitatively evaluate our method and its generative disentanglement features, see e.g., \citep{zhu2020s3vae, bai2021contrastively, naiman2023sample}. Specifically, we sample a sequence $x_{1:T}$ from the test set, and we disentangle it to static $s$ and dynamic $d_{1:T}$ factors. Then, we sample a new static code from the prior, which we denote by $\overline{s}$. Finally, we reconstruct the video sequence corresponding to $\overline{x}_{1:T} := \text{dec}(\overline{s}, d_{1:T})$. Naturally, we expect that the reconstruction will share the same dynamic features as the inputs, but will have a different static behavior, e.g., a different person with the same expression.


We estimate the new sequences quantitatively by using a pre-trained classifier. For every reconstructed $\overline{x}_{1:T}$, the classifier outputs its dynamic class, e.g., facial expression in MUG and action in Sprites. We measure the performance of our model with respect to the classification accuracy (Acc), the inception score (IS), intra-entropy $H(y|x)$ and inter-entropy $H(y)$. See App.~\ref{app:metrics} for more details on the metrics. We detail in Tab.~\ref{tab:gen_sprites_mug} the results our method obtains, and we compare it to recent SOTA approaches, evaluated on the Sprites and MUG datasets. Similarly to recent works, we achieve $100\%$ accuracy on Sprites, and strong measures on the other metrics (equivalent to SPYL). Further, we report new state-of-the-art results on the MUG dataset with an accuracy of $\mathbf{86.90 \%}$, IS$\,=\mathbf{5.598}$, $H(y|x) = \mathbf{0.041}$ and $H(y) = \mathbf{1.782}$.
In Tab.~\ref{tab:gen_sprites_mug_std} in the appendix, we present the above results with standard deviations; these results emphasize our method's statistical significance.

\begin{figure*}[ht]
    \centering
    \includegraphics[width=.75\linewidth]{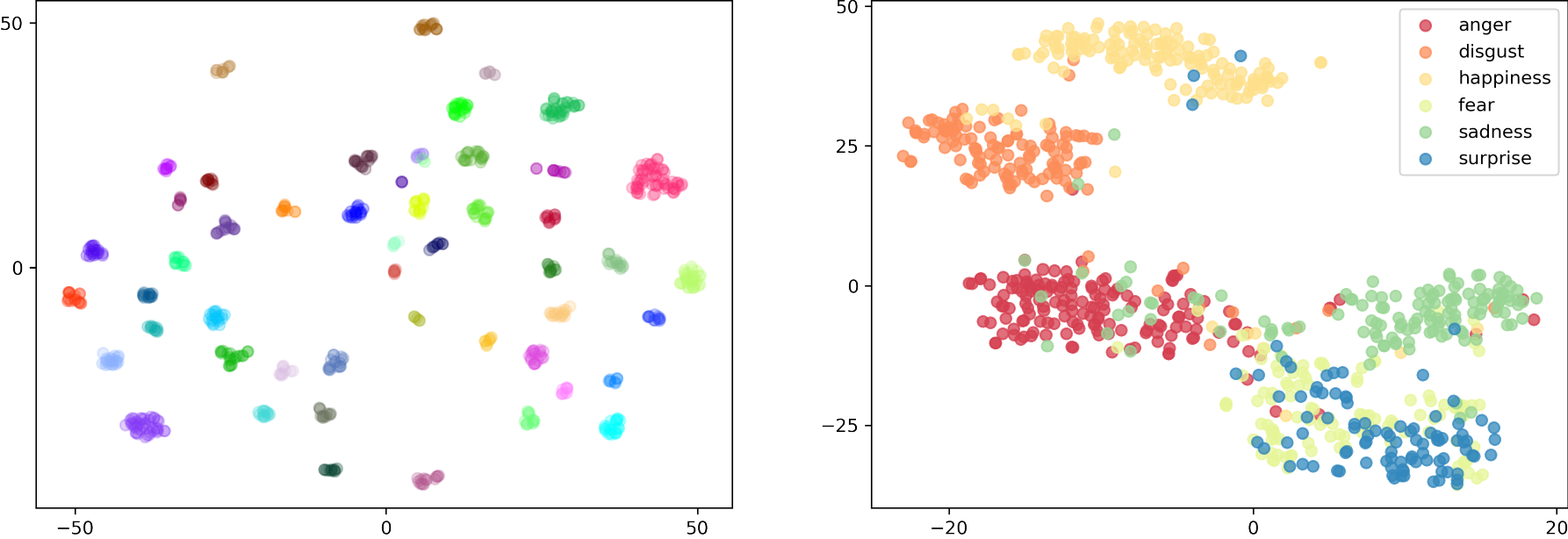}
    \vspace{-5mm}
    \caption{t-SNE plots on MUG dataset of the latent static and dynamic factors. Latent static codes, colored by subject identity (left), and latent dynamic codes, colored by dynamic attribute (right).} 
    \label{fig:tsne_mug}
\end{figure*}

\paragraph{Information leakage gap in video.} We extend the last experiment to show how our method mitigates information leakage. As mentioned above, the dynamics accuracy of $\overline{x}_{1:T}$ should be high. In contrast, the static category of this series should be random. However, if the static accuracy is higher than random, then some static information is encoded in $d_{1:T}$, i.e., there is information leakage. Additionally, we also perform the opposite test, namely, we sample a new dynamic code from the prior denoted by $\hat{d}_{1:T}$, and we reconstruct $\hat{x}_{1:T} := \text{dec}(s, \hat{d}_{1:T})$. In this case, we expect for high static accuracy and low dynamic classification. Testing both cases gives us a better understanding of the information leakage of each baseline approach. In Tab.~\ref{tab:full_dis_met_cls_task} (Generation), we show the accuracy for each test using the pre-trained classifier. We find that our approach yields high static and dynamic accuracy measures when sampling dynamic and static features, respectively, as expected. Notably, our method presents the best leakage gaps, obtaining $78.57\%$ and $84.01 \%$ improving SOTA by nearly $7\%$ and $2\%$, respectively. We also analyze in App.~\ref{app:failure_case_mug} the failure cases in MUG, showing that the fear and surprise are challenging to distinguish, and thus, even a $2\%$ improvement is significant for the dynamic gap.

We further extend our tests above, to highlight the latent encoding capabilities of our approach. We follow~\cite{naiman2023sample} and perform latent classification by extracting static codes $\{ s^j \}_{j=1}^N$, and split the resulting data into $80-20$ train-test sets. Then, we train two classifiers, where the first is trained to predict the static label of $s^j$, and the second is trained to predict the dynamics of $s^j$, for every $j$. Ideally, the first classifier would be able to predict the static features perfectly, and the second would fail to predict the corresponding dynamics, $d^j_{1:T}$.We also perform the complementing test where we train another two classifiers for $d^j_{1:T}$. Furthermore, we include in the experiment table a theoretical random guess baseline, illustrating the optimal outcome when the classifier is unable to predict the labels. In Tab.~\ref{tab:full_dis_met_cls_task} (Latent), we obtain a mitigation of information leakage where the gap over the static features is improved from $29.37\%$ (second best) to $54.29\%$ (ours), presenting a substantial improvement. In Tab.~\ref{tab:std_full_dis_met_cls_task} in the appendix, we show the above results with standard deviations, highlighting the significance of our method.

Furthermore, we consider more challenging datasets; we conduct experiments on the Jesters dataset \cite{materzynska2019jester}. This real-world complicated dataset of hundreds of subjects performing different hand gestures before a camera. This dataset is used in disentanglement tasks in \cite{naiman2023sample} as part of the latent classification benchmark (Tab.~\ref{tab:full_dis_met_cls_task} in our paper). We follow their benchmark protocol and evaluate our method according to it. We achieved significant improvement over the best current method. SPYL \cite{naiman2023sample} obtains a score of $27\%$ on the dynamic labels classification task with the dynamic features. At the same time, our method achieves more than two times better performance with a classification score of $68.2\% \pm 2.1\%$. Note that SPYL states that because only the hand gesture labels exist, only dynamic feature accuracy can be measured. These results illuminate our method's potential to perform well on large and complicated datasets.

\vspace{-5mm} 

\begin{figure*}[t]
    \centering
    \includegraphics[width=1\linewidth]{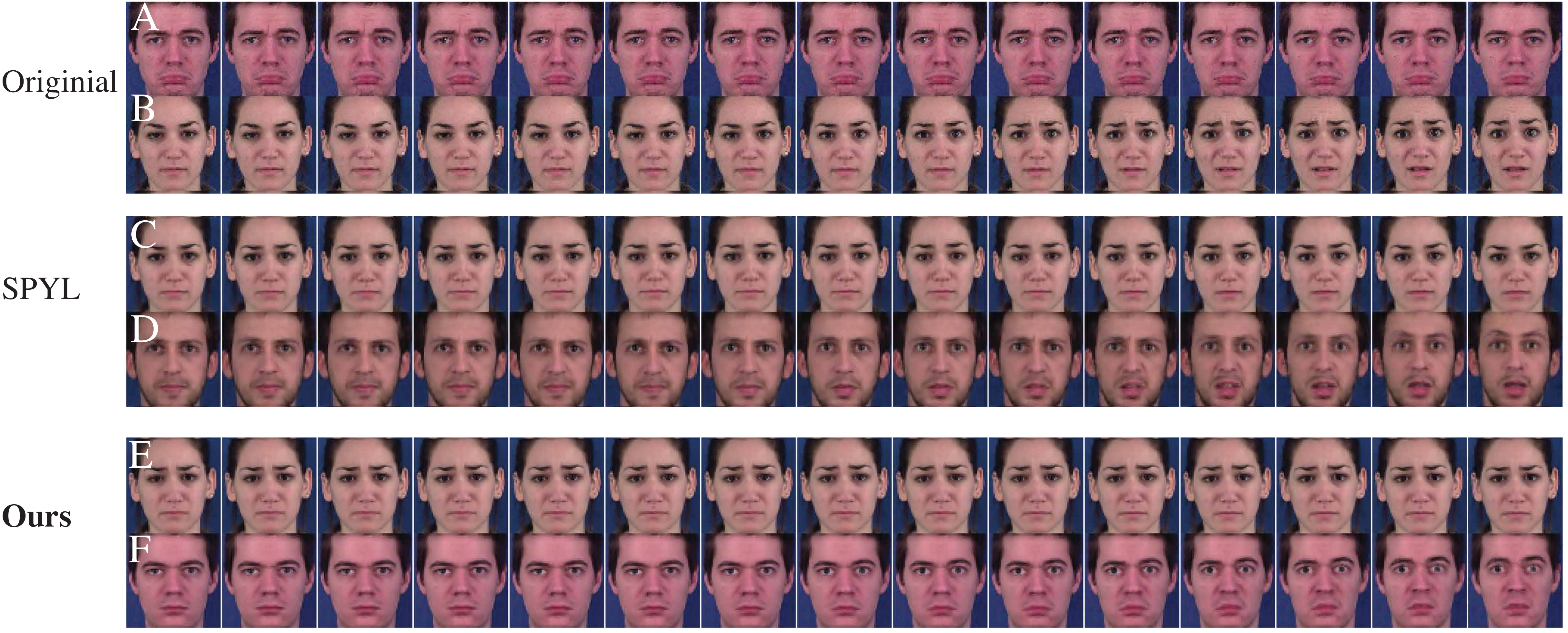}
    \vspace{-3mm}
    \caption{Qualitative example of swap between source and target sequences. A is the source, B is the target, C (E) is when static is swapped from source to target, and D (F) is when dynamics are swapped. C and D are the swaps for SPYL method and E and F are swaps of our method. We observe in D that the identity is changed when transferring the dynamics. See more in Sec.~\ref{subsec:qualitative_eval}, Fig.~\ref{fig:swap_comp_1} and Fig.~\ref{fig:swap_comp_2}.} 
    \label{fig:swap_mug_comp}
\end{figure*}

\paragraph{Image quality.} In addition to the disentanglement quality, we asses quantitatively the image quality of our model. We provide here the reconstruction losses (MSE) $224.35, 243.94$ and $223.36$ for C-DSVAE, SPYL, and our approach, respectively. These results indicate that our method achieves on par MSE in comparison to existing works, while significantly outperforming them on disentanglement tasks. In addition, we include a qualitative comparison of swaps and reconstructions in Figs.~\ref{fig:swap_mug}, \ref{fig:swap_mug_comp}.



\paragraph{Information leakage gap in audio.} The TIMIT dataset consists of different people that read different sentences. While every person reads unique sentences, distinguishing between speakers is possible by utilizing their static codes. We use the standard TIMIT benchmark for sequential disentanglement and report its results in App.~\ref{app:timit_experiment}. We obtain a static EER of $3.50\%$, dynamic EER of $34.62\%$, and disentanglement gap of $31.11\%$. Our results present a $\boldsymbol{1.3\%}$ gap improvement compared to current SOTA results. These results align with the above video experiments, reinforcing our model's ability to mitigate information leakage between latent codes also for audio modalities.

\subsubsection{Time Series Analysis}
\label{subsub:time_series_quant}

In what follows, we evaluate the effectiveness of our approach in disentangling static and dynamic features of time series information. We test the usefulness of the representations we extract for downstream tasks such as prediction and classification. This evaluation methodology aligns with previous work~\citep{oord2018representation, franceschi2019unsupervised, fortuin2020gp, tonekaboni2022decoupling}. Our results are compared with recent sequential disentanglement approaches and with techniques for time series data such as GP-VAE~\citep{fortuin2020gp} and GLR~\citep{tonekaboni2022decoupling}. For a fair comparison, we use the same encoder and decoder modules for all baseline methods.

\setlength{\tabcolsep}{3pt}
\begin{table}[!b]
    \centering
    \vspace{-5mm}
    \caption{Time series prediction benchmark.}
    \vskip 0.1in
    \label{tab:pred_ts_data}
    \begin{tabular}{lccc}
        \toprule
        & \multicolumn{2}{c}{PhysioNet} & \multicolumn{1}{c}{ETTh1} \\
        \textbf{Method} & \textbf{AUPRC} $\uparrow$ & \textbf{AUROC} $\uparrow$ & \textbf{MAE} $\downarrow$ \\
        \midrule
        VAE         & $0.157 \pm 0.05$ & $0.564 \pm 0.04$ & $13.66 \pm 0.20$    \\
        GP-VAE       & $0.282 \pm 0.09$ & $0.699 \pm 0.02$ & $14.98 \pm 0.41$    \\
        C-DSVAE     & $0.158 \pm 0.01$ & $0.565 \pm 0.01$ & $12.53 \pm 0.88$     \\
        GLR         & $0.365 \pm 0.09$ & $0.752 \pm 0.01$ & $12.27 \pm 0.03$    \\
        SPYL        & $0.367 \pm 0.02$ & $0.764 \pm 0.04$ & $12.22 \pm 0.03$     \\
        \midrule
        Ours w.o. loss & $0.274 \pm 0.02$ & $0.692 \pm 0.01$ & $18.32 \pm 0.32$ \\
        Ours w.o. sub  & $0.411 \pm 0.02$ & $0.798 \pm 0.01$ & $16.53 \pm 0.13$ \\
        Ours w.o. both & $0.255 \pm 0.02$ & $0.631 \pm 0.02$ & $17.42 \pm 0.04$ \\
        \textbf{Ours} & \textbf{0.473 $\pm$ 0.02} & \textbf{0.858 $\pm$ 0.01} &  \textbf{11.21 $\pm$ 0.01} \\
        \midrule
        RF  & $0.446 \pm 0.04$ & $0.802 \pm 0.04$ &  $10.19 \pm 0.20$    \\
        \bottomrule
    \end{tabular}
    \vskip -0.1in
\end{table}

\paragraph{Downstream prediction tasks.} We consider two cases: (i) predicting the risk of in-hospital mortality using the PhysioNet dataset; and (ii) predicting the oil temperature of electricity transformer with ETTh1. In both cases, we train our model on sequences $x_{1:T}$ to learn disentangled static and dynamics features, denoted by $s$ and $d_{1:T}$, respectively. Then, we use the extracted codes as the train set of a simple predictor network. Please see further details in App.~\ref{app:hyperparameters}. We compute the AUROC and AUPRC error measures on PhysioNet, and the mean absolute error (MAE) for ETTh1, and we report the results in Tab.~\ref{tab:pred_ts_data}. We also include the baseline results for training directly on the raw features, appearing as `RF` in the table. Remarkably, our method achieves SOTA results and it outperforms all other baseline approaches on the mortality prediction task, including the baseline that trains on the original data. Thus, our results highlight that unlike previous works, our disentangled representations effectively improve downstream prediction tasks--a highly-sought characteristic in representation learning \citep{bengio2013representation, tonekaboni2022decoupling}.


\paragraph{Static patterns identification.} Reducing prediction errors could be potentially achieved by considering global patterns \cite{trivedi2015utility}, however, the given data may include non-useful features. Here, we explore if our model can extract global patterns through the static features. We follow~\cite{tonekaboni2022decoupling} and select the ICU unit as the global label for PhysioNet and month of the year for the Air Quality dataset. We train our model and compute static features, followed by training a simple MLP for classifying the extracted codes. The classification results are summarized in Tab.~\ref{tab:classify_ts_data}, where we find that our model better encodes global patterns in $s$ in comparison to baseline techniques.

\subsection{Qualitative Evaluation}
\label{subsec:qualitative_eval}

We qualitatively assess our approach on video sequences and general time series data. To this end, we analyze the learned latent space of static and dynamic factors by utilizing t-SNE~\citep{van2008visualizing}. In addition, we also swap factors of variation between two separate sequences, similar to e.g., \citet{li2018disentangled}.

\paragraph{Static and dynamic clustering.} To assess the capacity of our approach to disentangle data into distinct subspaces, we perform the following: First, we randomly select a batch of samples from a given test set. Second, we extract the static $s^j$ and dynamic $d_{1:T}^j$ latent disentangled representations for each sample $j$ from the batch. Third, we compute $d^j := \sum_t d^j_t / T$. Thus, $d^j \in \mathbb{R}^k$ is the averaged dynamic sequence across time. The latent dimension $k$ is a user parameter, and we set it to match the dimension of the static code to emphasize that in our approach, these vectors could be of the same dimension while obtaining good disentanglement, unlike competing works such as DSVAE. Given $d^j$ and $s$, we compute their t-SNE embedding, where we observed that computing it separately or together did not change the results. We visualize the obtained embedding of MUG dataset in Fig.~\ref{fig:tsne_mug}, where static factors are colored per subject and the dynamic factors per expression. The results show that our approach clearly clusters between the time-varying and time-invariant factors of variation. Further, we can observe distinct static sub-clusters, which may indicate a hierarchical clustering based on the identities of subjects. Indeed, our model learns a clustered representation with respect to people's identities without any explicit constraints. In addition, we plot the dynamic embeddings colored by their labels. We observe a clustered representation for most of the attributes. The 'fear' and 'surprise' dynamics representations seem to overlap, and we elaborate on this in App.~\ref{app:failure_case_mug}.

\begin{figure}[!hb]
    \centering
    \includegraphics[width=1\linewidth]{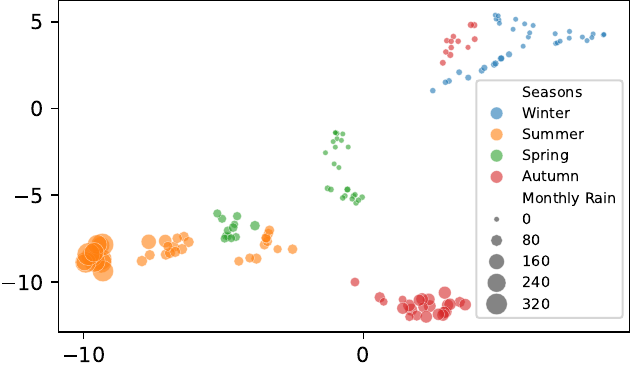}
    \vspace{-5mm}
    \caption{t-SNE visualization of static features from the Air Quality dataset, depicting the model's ability to distinguish days based on precipitation, irrespective of season. Each point represents a day, colored by season and scaled by the amount of rain, illustrating the model's nuanced clustering of dry and wet days within the static seasonal context.}
    \label{fig:tsne_time_series_rain}
\end{figure}

Moreover, we also embed the static codes of Air Quality in Beijing using t-SNE in Fig.~\ref{fig:tsne_time_series_rain}. We scale each point by the amount of rainfall corresponding to that day, and we color the points by the related season. Our results show a clear clustering of the latent codes with respect to the rainy seasons. That is, the Summer and its adjacent days from Spring and Autumn are clustered together as opposed to the Winter, which is very dry in Beijing. 

\begin{figure*}[t]
    \centering
    \begin{overpic}[width=1\linewidth]{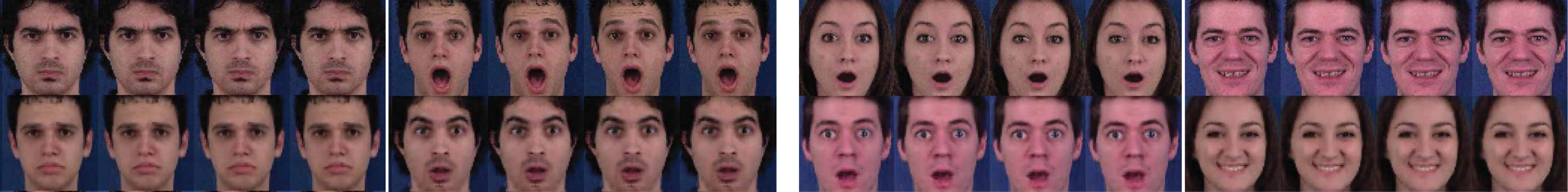}
        \put(0,11){\scriptsize \color{white}A} \put(51.25,11){\scriptsize \color{white}A} 
        \put(25,11){\scriptsize \color{white}B} \put(75.75,11){\scriptsize \color{white}B}
        \put(0,5){\scriptsize \color{white}C} \put(51.25,5){\scriptsize \color{white}C}
        \put(25,5){\scriptsize \color{white}D} \put(75.75,5){\scriptsize \color{white}D}
    \end{overpic}
    \vspace{-3mm}
    \caption{Two qualitative examples of swap between source and target sequences. A is the source, B is the target, C is when static is swapped from source to target, and D is when dynamics are swapped. See details in Sec.~\ref{subsec:qualitative_eval}.} 
    \label{fig:swap_mug}
\end{figure*}

\paragraph{Sequential swap.} We will now show the ability of our model to \emph{swap} between static and dynamic representations of two different sequences. Specifically, we take $x^{1}_{1:T}, x^{2}_{1:T}$ from a given test set, and we extract their static and dynamic codes, $s^1, s^2$ and $d^{1}_{1:T}, d^{2}_{1:T}$. Then, we swap the representations by joining the static code of the first sequence with the dynamic factors of the second sequence, and vice versa. Formally, we generate new samples $\overline{x}^1_{1:T}$ and $\overline{x}^2_{1:T}$ via $\overline{x}^1_{1:T} := \text{dec}(s^1 \ , d^2_{1:T})$, and $\overline{x}^2_{1:T} := \text{dec}(s^2 \ , d^1_{1:T})$, where $\text{dec}$ is the decoder. In Fig.~\ref{fig:swap_mug}, we show two swap examples for the MUG dataset. Additional examples for Sprites and MUG are shown in Figs.~\ref{fig:sprites_swap_content}, \ref{fig:swap_mug_1}. 

\setlength{\tabcolsep}{5pt}
\begin{table}[!hb]
    \centering
    \caption{Time series classification benchmark.}
    \vskip 0.1in
    \label{tab:classify_ts_data}
    \begin{tabular}{lcc}
        \toprule
        \multicolumn{2}{c}{} \\
        Method & PhysioNet $\uparrow$ & Air Quality $\uparrow$ \\
        \midrule
        VAE        & $34.71 \pm 0.23$ & $27.17 \pm 0.03$ \\
        GP-VAE     & $42.47 \pm 2.02$ & $36.73 \pm 1.40$ \\
        C-DSVAE    & $32.54 \pm 0.00$ & $47.07 \pm 1.20$  \\
        GLR        & $38.93 \pm 2.48$ & $50.32 \pm 3.87$  \\
        SPYL       & $46.98 \pm 3.04$ & $57.93 \pm 3.53$  \\
        \midrule
        Ours w.o. loss  & $42.16 \pm 0.104$ & $50.14 \pm 0.013$ \\
        Ours w.o. sub   & $46.15 \pm 0.014$ & $56.11 \pm 0.021$  \\
        Ours w.o. both  & $41.42 \pm 0.019$ & $48.32 \pm 0.102$ \\
        Ours            & $\bm{56.87 \pm 0.34}$ & $\bm{65.87 \pm 0.01}$  \\
        \midrule
        RF         & $62.00 \pm 2.10$ & $62.43 \pm 0.54$  \\
        \bottomrule
    \end{tabular}
    \vskip -0.1in
\end{table}

\subsection{Ablation Studies}
\label{subsec:ablt_studies}
\paragraph{Objective function and architecture.} We perform an ablation study to justify our split loss term and our architectural choice of the subtraction module. This study comprises three specific variations of training our method: (1) without the static loss in Eq.~\ref{eq:our_recon_loss} (no loss), (2) without subtracting the static representation from the learned dynamics (no sub), and (3) without both of them (no both). In all these cases, we kept the hyper-parameters constant. We show the ablation results in Tab.~\ref{tab:full_dis_met_cls_task} for MUG, and in Tab.~\ref{tab:pred_ts_data} for PhysioNet and ETTh1. Evidently, the removal of components negatively impacts the disentanglement capabilities of our method across multiple tasks. For instance, we observe a significant information leakage on the MUG dataset, illustrated in low accuracy results for the ablation models, e.g., performance drops by $\approx 46\%$ and $\approx 65\%$ for the no loss and no sub baselines, respectively. Similarly to the MUG case, we also identify a significant performance decrease in the time series tasks.

\setlength{\tabcolsep}{3pt}
\begin{table}[ht]
    \centering
    \caption{Ablation study of our model components (top) and its robustness to index choice (bottom).}
    \vskip 0.1in
    \label{tab:ablation_ts_data}
    \begin{tabular}{lccc}
        \toprule
        & \multicolumn{2}{c}{PhysioNet} & \multicolumn{1}{c}{ETTh1} \\
        \textbf{Method} & \textbf{AUPRC} $\uparrow$ & \textbf{AUROC} $\uparrow$ & \textbf{MAE} $\downarrow$ \\
        \midrule
        $x_\text{rob}$ & $0.276 \pm 0.013$ & $0.501 \pm 0.033$ & $13.56 \pm 0.051$ \\
        $x_\text{random}$ & $0.425 \pm 0.006$ & $0.837 \pm 0.018$ & $11.26 \pm 0.011$ \\
        $x_{T/2}$ & $0.434 \pm 0.001$ & $0.842 \pm 0.016$ & $11.32 \pm 0.014$ \\
        $x_T$  & $0.441 \pm 0.009$ & $0.839 \pm 0.002$ & $11.11 \pm 0.006$ \\
        \midrule
        $\boldsymbol{x_1}$  & $0.473 \pm 0.021$ & $0.858 \pm 0.006$ &  $11.21 \pm 0.001$    \\
        \bottomrule
    \end{tabular}
    \vskip -0.1in
\end{table}

\paragraph{Dependence on the $i$-th sample.} One potential shortcoming of our approach is its dependence on extracting static information from the $i$-th sample. Here, we would like to empirically verify the robustness of our method to the choice of sample. Thus, instead of using the $i$-th sample, we also consider taking the middle $x_{T/2}$ and last $x_T$ samples, i.e., we train models learning the distributions $\dist{q}{s}{x_{T/2} \, ; \, \phi_s}$ and $\dist{q}{s}{x_T \, ; \, \phi_s}$. We further extend our evaluation of the dependence on the $i$ sample experiment. Instead of picking $i$ from $1, T/2, T$, we \textit{randomly} sample $i$ and run the experiment. To ensure variability in the choice of $i$ and to demonstrate the robustness of our method, we repeat the experiment \textit{six} times with randomly picked $i$. We report the average result across all runs and the standard deviation under the method name $x_{\text{random}}$. Additionally, we emphasize that randomly picking samples from the sequence during training for each batch leads to information leakage. This training process "exposes" the static module that is responsible for extracting static information to the different samples from the same sequence. Therefore, dynamic information can leak into the static representation. We add the results of this experiment under the method name $x_\text{rob}$ (\textbf{r}andom \textbf{o}n \textbf{b}atch) to demonstrate this phenomenon. We present in Tab.~\ref{tab:ablation_ts_data}, Tab.~\ref{tab:ablation_ts_data_app_t}, and Tab.~\ref{tab:ablation_frame_mug} the ablation results on several datasets and tasks. Our results indicate that selecting other samples does not impact the results significantly, highlighting the robustness of our method to the choice of sample.

\section{Limitations and Conclusion}


There are two limitations of our approach. The first limitation is the dependency on the VAE model and it is shared by most existing works, often leading to low quality reconstructed and generated samples. The second shortcoming is unique to our approach and it is related to the dependency on a single element for extracting the static information. While this inductive bias may seem too limiting, our extensive evaluation and ablation results show that the choice of sample we use has little effect on the model behavior and overall performance over the standard benchmark. Similar to previous works, our model assumes that the static factor exists in every element of the sequence. With a glance to the future, our new inductive bias might be limiting in settings where there are sequence elements that do not include any static information. We emphasize that the current sequential disentanglement literature does not deal with this scenario at all, since all models assume that the static factor is shared across the entire sequence. We leave further exploration and additional investigation of this aspect for future work.

In this work, we considered the sequential disentanglement challenge, in which input sequences are decomposed to a single time-invariant component and a series of time-varying factors of variation. Unfortunately, existing variational models suffer from information leakage issues, where the dynamic factors encode all of the information in the data, and the static code remains non-meaningful. Resolving information leakage issues via changing the latent subspace dimension and by incorporating mutual information loss terms have been showing limited success. In our work, we observe that the static code may be extracted from a single sample, yielding a new posterior variational model. Further, we design a new neural network that alleviates information leakage issues by subtracting the extracted static code from the learned dynamic factors. The resulting model is easy-to-code and it has less hyper-parameters in comparison to state-of-the-art approaches. We extensively evaluate our method on standard sequential disentanglement benchmarks including general time series, video and audio datasets. Our model outperforms existing work on generation and prediction tasks as measured by various qualitative and quantitative metrics. In the future, we would like to investigate other architectural backbones such as diffusion models, and we would like to explore sequential disentanglement on real-world datasets and problems. 

\section*{Impact Statement}

Extracting static and dynamic features creates opportunities to interchange between identities and actions. This capability could be exploited for malicious purposes, such as the creation of fake news or identity theft. While our disentanglement approach may not allow for the seamless switching of intricate actions and identities, it is crucial to broaden the discussion regarding the ethical implications associated with the development and application of these methods.

\section*{Acknowledgements}
This research was partially supported by the Lynn and
William Frankel Center of the Computer Science Department, Ben-Gurion University of the Negev, an ISF grant
668/21, an ISF equipment grant, and by the Israeli Council
for Higher Education (CHE) via the Data Science Research
Center, Ben-Gurion University of the Negev, Israel.

\clearpage
\bibliographystyle{icml2024}

\begin{thebibliography}{49}
\providecommand{\natexlab}[1]{#1}
\providecommand{\url}[1]{\texttt{#1}}
\expandafter\ifx\csname urlstyle\endcsname\relax
  \providecommand{\doi}[1]{doi: #1}\else
  \providecommand{\doi}{doi: \begingroup \urlstyle{rm}\Url}\fi

\bibitem[Aifanti et~al.(2010)Aifanti, Papachristou, and Delopoulos]{aifanti2010mug}
Aifanti, N., Papachristou, C., and Delopoulos, A.
\newblock The {MUG} facial expression database.
\newblock In \emph{11th International Workshop on Image Analysis for Multimedia Interactive Services WIAMIS 10}, pp.\  1--4. IEEE, 2010.

\bibitem[Bahdanau et~al.(2015)Bahdanau, Cho, and Bengio]{bahdanau2014neural}
Bahdanau, D., Cho, K., and Bengio, Y.
\newblock Neural machine translation by jointly learning to align and translate.
\newblock In \emph{3rd International Conference on Learning Representations, {ICLR}}, 2015.

\bibitem[Bai et~al.(2021)Bai, Wang, and Gomes]{bai2021contrastively}
Bai, J., Wang, W., and Gomes, C.~P.
\newblock Contrastively disentangled sequential variational autoencoder.
\newblock \emph{Advances in Neural Information Processing Systems}, 34:\penalty0 10105--10118, 2021.

\bibitem[Bengio et~al.(2013)Bengio, Courville, and Vincent]{bengio2013representation}
Bengio, Y., Courville, A., and Vincent, P.
\newblock Representation learning: a review and new perspectives.
\newblock \emph{IEEE transactions on pattern analysis and machine intelligence}, 35\penalty0 (8):\penalty0 1798--1828, 2013.

\bibitem[Berman et~al.(2023)Berman, Naiman, and Azencot]{berman2023multifactor}
Berman, N., Naiman, I., and Azencot, O.
\newblock Multifactor sequential disentanglement via structured koopman autoencoders.
\newblock In \emph{The Eleventh International Conference on Learning Representations, {ICLR}}, 2023.

\bibitem[Bhagat et~al.(2020)Bhagat, Uppal, Yin, and Lim]{bhagat2020disentangling}
Bhagat, S., Uppal, S., Yin, Z., and Lim, N.
\newblock Disentangling multiple features in video sequences using gaussian processes in variational autoencoders.
\newblock In \emph{Computer Vision--ECCV 2020: 16th European Conference, Glasgow, UK, August 23--28, 2020, Proceedings, Part XXIII 16}, pp.\  102--117. Springer, 2020.

\bibitem[Bouchacourt et~al.(2018)Bouchacourt, Tomioka, and Nowozin]{bouchacourt2018multi}
Bouchacourt, D., Tomioka, R., and Nowozin, S.
\newblock Multi-level variational autoencoder: learning disentangled representations from grouped observations.
\newblock In \emph{Proceedings of the AAAI Conference on Artificial Intelligence}, volume~32, 2018.

\bibitem[Chen et~al.(2018)Chen, Li, Grosse, and Duvenaud]{chen2018isolating}
Chen, R.~T., Li, X., Grosse, R.~B., and Duvenaud, D.~K.
\newblock Isolating sources of disentanglement in variational autoencoders.
\newblock \emph{Advances in neural information processing systems}, 31, 2018.

\bibitem[Chen et~al.(2016)Chen, Duan, Houthooft, Schulman, Sutskever, and Abbeel]{chen2016infogan}
Chen, X., Duan, Y., Houthooft, R., Schulman, J., Sutskever, I., and Abbeel, P.
\newblock {InfoGAN:} interpretable representation learning by information maximizing generative adversarial nets.
\newblock \emph{Advances in neural information processing systems}, 29, 2016.

\bibitem[Chen et~al.(2017)Chen, Kingma, Salimans, Duan, Dhariwal, Schulman, Sutskever, and Abbeel]{chen2016variational}
Chen, X., Kingma, D.~P., Salimans, T., Duan, Y., Dhariwal, P., Schulman, J., Sutskever, I., and Abbeel, P.
\newblock Variational lossy autoencoder.
\newblock In \emph{5th International Conference on Learning Representations, {ICLR}}, 2017.

\bibitem[Chenafa et~al.(2008)Chenafa, Istrate, Vrabie, and Herbin]{chenafa2008biometric}
Chenafa, M., Istrate, D., Vrabie, V., and Herbin, M.
\newblock Biometric system based on voice recognition using multiclassifiers.
\newblock In \emph{Biometrics and Identity Management: First European Workshop, BIOID 2008, Roskilde, Denmark, May 7-9, 2008. Revised Selected Papers 1}, pp.\  206--215. Springer, 2008.

\bibitem[Chung et~al.(2015)Chung, Kastner, Dinh, Goel, Courville, and Bengio]{chung2015recurrent}
Chung, J., Kastner, K., Dinh, L., Goel, K., Courville, A.~C., and Bengio, Y.
\newblock A recurrent latent variable model for sequential data.
\newblock \emph{Advances in neural information processing systems}, 28, 2015.

\bibitem[Denton \& Birodkar(2017)Denton and Birodkar]{denton2017unsupervised}
Denton, E.~L. and Birodkar, V.
\newblock Unsupervised learning of disentangled representations from video.
\newblock \emph{Advances in neural information processing systems}, 30, 2017.

\bibitem[Erichson et~al.(2021)Erichson, Azencot, Queiruga, Hodgkinson, and Mahoney]{erichson2021lipschitz}
Erichson, N.~B., Azencot, O., Queiruga, A.~F., Hodgkinson, L., and Mahoney, M.~W.
\newblock Lipschitz recurrent neural networks.
\newblock In \emph{9th International Conference on Learning Representations, {ICLR}}, 2021.

\bibitem[Fortuin et~al.(2020)Fortuin, Baranchuk, R{\"a}tsch, and Mandt]{fortuin2020gp}
Fortuin, V., Baranchuk, D., R{\"a}tsch, G., and Mandt, S.
\newblock {GP-VAE}: deep probabilistic time series imputation.
\newblock In \emph{International conference on artificial intelligence and statistics}, pp.\  1651--1661. PMLR, 2020.

\bibitem[Franceschi et~al.(2019)Franceschi, Dieuleveut, and Jaggi]{franceschi2019unsupervised}
Franceschi, J.-Y., Dieuleveut, A., and Jaggi, M.
\newblock Unsupervised scalable representation learning for multivariate time series.
\newblock \emph{Advances in neural information processing systems}, 32, 2019.

\bibitem[Garofolo(1993)]{garofolo1993timit}
Garofolo, J.~S.
\newblock {TIMIT} acoustic phonetic continuous speech corpus.
\newblock \emph{Linguistic Data Consortium, 1993}, 1993.

\bibitem[Girin et~al.(2021)Girin, Leglaive, Bie, Diard, Hueber, and Alameda{-}Pineda]{girin2021dynamical}
Girin, L., Leglaive, S., Bie, X., Diard, J., Hueber, T., and Alameda{-}Pineda, X.
\newblock Dynamical variational autoencoders: {A} comprehensive review.
\newblock \emph{Found. Trends Mach. Learn.}, 15\penalty0 (1-2):\penalty0 1--175, 2021.

\bibitem[Goldberger et~al.(2000)Goldberger, Amaral, Glass, Hausdorff, Ivanov, Mark, Mietus, Moody, Peng, and Stanley]{goldberger2000physiobank}
Goldberger, A.~L., Amaral, L.~A., Glass, L., Hausdorff, J.~M., Ivanov, P.~C., Mark, R.~G., Mietus, J.~E., Moody, G.~B., Peng, C.-K., and Stanley, H.~E.
\newblock {PhysioBank}, {PhysioToolkit}, and {PhysioNet}: components of a new research resource for complex physiologic signals.
\newblock \emph{circulation}, 101\penalty0 (23):\penalty0 e215--e220, 2000.

\bibitem[Han et~al.(2021)Han, Min, Han, Li, and Zhang]{han2021disentangled}
Han, J., Min, M.~R., Han, L., Li, L.~E., and Zhang, X.
\newblock Disentangled recurrent {Wasserstein} autoencoder.
\newblock In \emph{9th International Conference on Learning Representations, {ICLR}}, 2021.

\bibitem[Higgins et~al.(2016)Higgins, Matthey, Pal, Burgess, Glorot, Botvinick, Mohamed, and Lerchner]{higgins2016beta}
Higgins, I., Matthey, L., Pal, A., Burgess, C., Glorot, X., Botvinick, M., Mohamed, S., and Lerchner, A.
\newblock beta-{VAE}: learning basic visual concepts with a constrained variational framework.
\newblock In \emph{International conference on learning representations}, 2016.

\bibitem[Hochreiter \& Schmidhuber(1997)Hochreiter and Schmidhuber]{hochreiter1997long}
Hochreiter, S. and Schmidhuber, J.
\newblock Long short-term memory.
\newblock \emph{Neural computation}, 9\penalty0 (8):\penalty0 1735--1780, 1997.

\bibitem[Hsieh et~al.(2018)Hsieh, Liu, Huang, Fei-Fei, and Niebles]{hsieh2018learning}
Hsieh, J.-T., Liu, B., Huang, D.-A., Fei-Fei, L.~F., and Niebles, J.~C.
\newblock Learning to decompose and disentangle representations for video prediction.
\newblock \emph{Advances in neural information processing systems}, 31, 2018.

\bibitem[Hsu et~al.(2017)Hsu, Zhang, and Glass]{hsu2017unsupervised}
Hsu, W.-N., Zhang, Y., and Glass, J.
\newblock Unsupervised learning of disentangled and interpretable representations from sequential data.
\newblock \emph{Advances in neural information processing systems}, 30, 2017.

\bibitem[Kim \& Mnih(2018)Kim and Mnih]{kim2018disentangling}
Kim, H. and Mnih, A.
\newblock Disentangling by factorising.
\newblock In \emph{International Conference on Machine Learning}, pp.\  2649--2658. PMLR, 2018.

\bibitem[Kingma \& Welling(2014)Kingma and Welling]{kingma14auto}
Kingma, D.~P. and Welling, M.
\newblock Auto-encoding variational bayes.
\newblock In \emph{2nd International Conference on Learning Representations, {ICLR}}, 2014.

\bibitem[Kumar et~al.(2018)Kumar, Sattigeri, and Balakrishnan]{kumar2017variational}
Kumar, A., Sattigeri, P., and Balakrishnan, A.
\newblock Variational inference of disentangled latent concepts from unlabeled observations.
\newblock In \emph{6th International Conference on Learning Representations, {ICLR}}, 2018.

\bibitem[LeCun et~al.(1989)LeCun, Boser, Denker, Henderson, Howard, Hubbard, and Jackel]{lecun1989backpropagation}
LeCun, Y., Boser, B., Denker, J.~S., Henderson, D., Howard, R.~E., Hubbard, W., and Jackel, L.~D.
\newblock Backpropagation applied to handwritten zip code recognition.
\newblock \emph{Neural computation}, 1\penalty0 (4):\penalty0 541--551, 1989.

\bibitem[Li \& Mandt(2018)Li and Mandt]{li2018disentangled}
Li, Y. and Mandt, S.
\newblock Disentangled sequential autoencoder.
\newblock In \emph{Proceedings of the 35th International Conference on Machine Learning, {ICML}}, volume~80, pp.\  5656--5665, 2018.

\bibitem[Locatello et~al.(2020)Locatello, Tschannen, Bauer, R{\"{a}}tsch, Sch{\"{o}}lkopf, and Bachem]{locatello2020disentangling}
Locatello, F., Tschannen, M., Bauer, S., R{\"{a}}tsch, G., Sch{\"{o}}lkopf, B., and Bachem, O.
\newblock Disentangling factors of variations using few labels.
\newblock In \emph{8th International Conference on Learning Representations, {ICLR}}, 2020.

\bibitem[Materzynska et~al.(2019)Materzynska, Berger, Bax, and Memisevic]{materzynska2019jester}
Materzynska, J., Berger, G., Bax, I., and Memisevic, R.
\newblock The jester dataset: A large-scale video dataset of human gestures.
\newblock In \emph{Proceedings of the IEEE/CVF international conference on computer vision workshops}, pp.\  0--0, 2019.

\bibitem[Naiman \& Azencot(2023)Naiman and Azencot]{naiman2023operator}
Naiman, I. and Azencot, O.
\newblock An operator theoretic approach for analyzing sequence neural networks.
\newblock In \emph{Proceedings of the AAAI conference on artificial intelligence}, volume~37, pp.\  9268--9276, 2023.

\bibitem[Naiman et~al.(2023)Naiman, Berman, and Azencot]{naiman2023sample}
Naiman, I., Berman, N., and Azencot, O.
\newblock Sample and predict your latent: modality-free sequential disentanglement via contrastive estimation.
\newblock In \emph{International Conference on Machine Learning, {ICML}}, volume 202, pp.\  25694--25717, 2023.

\bibitem[Naiman et~al.(2024)Naiman, Erichson, Ren, Mahoney, and Azencot]{naiman2024generative}
Naiman, I., Erichson, N.~B., Ren, P., Mahoney, M.~W., and Azencot, O.
\newblock Generative modeling of regular and irregular time series data via koopman vaes.
\newblock In \emph{The Eleventh International Conference on Learning Representations, {ICLR}}, 2024.

\bibitem[Oord et~al.(2018)Oord, Li, and Vinyals]{oord2018representation}
Oord, A. v.~d., Li, Y., and Vinyals, O.
\newblock Representation learning with contrastive predictive coding.
\newblock \emph{arXiv preprint arXiv:1807.03748}, 2018.

\bibitem[Reed et~al.(2015)Reed, Zhang, Zhang, and Lee]{reed2015deep}
Reed, S.~E., Zhang, Y., Zhang, Y., and Lee, H.
\newblock Deep visual analogy-making.
\newblock \emph{Advances in neural information processing systems}, 28, 2015.

\bibitem[Sch{\"o}lkopf et~al.(2021)Sch{\"o}lkopf, Locatello, Bauer, Ke, Kalchbrenner, Goyal, and Bengio]{scholkopf2021toward}
Sch{\"o}lkopf, B., Locatello, F., Bauer, S., Ke, N.~R., Kalchbrenner, N., Goyal, A., and Bengio, Y.
\newblock Toward causal representation learning.
\newblock \emph{Proceedings of the IEEE}, 109\penalty0 (5):\penalty0 612--634, 2021.

\bibitem[Tian et~al.(2021)Tian, Ren, Chai, Olszewski, Peng, Metaxas, and Tulyakov]{tian2021a}
Tian, Y., Ren, J., Chai, M., Olszewski, K., Peng, X., Metaxas, D.~N., and Tulyakov, S.
\newblock A good image generator is what you need for high-resolution video synthesis.
\newblock In \emph{9th International Conference on Learning Representations, {ICLR}}, 2021.

\bibitem[Tonekaboni et~al.(2022)Tonekaboni, Li, Arik, Goldenberg, and Pfister]{tonekaboni2022decoupling}
Tonekaboni, S., Li, C.-L., Arik, S.~O., Goldenberg, A., and Pfister, T.
\newblock Decoupling local and global representations of time series.
\newblock In \emph{International Conference on Artificial Intelligence and Statistics}, pp.\  8700--8714. PMLR, 2022.

\bibitem[Trivedi et~al.(2015)Trivedi, Pardos, and Heffernan]{trivedi2015utility}
Trivedi, S., Pardos, Z.~A., and Heffernan, N.~T.
\newblock The utility of clustering in prediction tasks.
\newblock \emph{arXiv preprint arXiv:1509.06163}, 2015.

\bibitem[Tulyakov et~al.(2018)Tulyakov, Liu, Yang, and Kautz]{tulyakov2018mocogan}
Tulyakov, S., Liu, M.-Y., Yang, X., and Kautz, J.
\newblock {MoCoGAN}: decomposing motion and content for video generation.
\newblock In \emph{Proceedings of the IEEE conference on computer vision and pattern recognition}, pp.\  1526--1535, 2018.

\bibitem[Van~der Maaten \& Hinton(2008)Van~der Maaten and Hinton]{van2008visualizing}
Van~der Maaten, L. and Hinton, G.
\newblock Visualizing data using {t-SNE}.
\newblock \emph{Journal of machine learning research}, 9\penalty0 (11), 2008.

\bibitem[Villegas et~al.(2017)Villegas, Yang, Hong, Lin, and Lee]{villegas2017decomposing}
Villegas, R., Yang, J., Hong, S., Lin, X., and Lee, H.
\newblock Decomposing motion and content for natural video sequence prediction.
\newblock In \emph{5th International Conference on Learning Representations, {ICLR}}, 2017.

\bibitem[Wu et~al.(2021)Wu, Lischinski, and Shechtman]{wu2021stylespace}
Wu, Z., Lischinski, D., and Shechtman, E.
\newblock Stylespace analysis: disentangled controls for {StyleGAN} image generation.
\newblock In \emph{Proceedings of the IEEE/CVF Conference on Computer Vision and Pattern Recognition}, pp.\  12863--12872, 2021.

\bibitem[Yamada et~al.(2020)Yamada, Kim, Miyoshi, Iwata, and Yamakawa]{yamada2020disentangled}
Yamada, M., Kim, H., Miyoshi, K., Iwata, T., and Yamakawa, H.
\newblock Disentangled representations for sequence data using information bottleneck principle.
\newblock In \emph{Asian Conference on Machine Learning}, pp.\  305--320. PMLR, 2020.

\bibitem[Zhang et~al.(2017)Zhang, Guo, Dong, He, Xu, and Chen]{zhang2017cautionary}
Zhang, S., Guo, B., Dong, A., He, J., Xu, Z., and Chen, S.~X.
\newblock Cautionary tales on air-quality improvement in {Beijing}.
\newblock \emph{Proceedings of the Royal Society A: Mathematical, Physical and Engineering Sciences}, 473\penalty0 (2205):\penalty0 20170457, 2017.

\bibitem[Zhao et~al.(2019)Zhao, Song, and Ermon]{zhao2019infovae}
Zhao, S., Song, J., and Ermon, S.
\newblock Infovae: Balancing learning and inference in variational autoencoders.
\newblock In \emph{Proceedings of the aaai conference on artificial intelligence}, pp.\  5885--5892, 2019.

\bibitem[Zhou et~al.(2021)Zhou, Zhang, Peng, Zhang, Li, Xiong, and Zhang]{zhou2021informer}
Zhou, H., Zhang, S., Peng, J., Zhang, S., Li, J., Xiong, H., and Zhang, W.
\newblock Informer: beyond efficient transformer for long sequence time-series forecasting.
\newblock In \emph{Proceedings of the AAAI conference on artificial intelligence}, pp.\  11106--11115, 2021.

\bibitem[Zhu et~al.(2020)Zhu, Min, Kadav, and Graf]{zhu2020s3vae}
Zhu, Y., Min, M.~R., Kadav, A., and Graf, H.~P.
\newblock {S3VAE}: self-supervised sequential {VAE} for representation disentanglement and data generation.
\newblock In \emph{Proceedings of the IEEE/CVF Conference on Computer Vision and Pattern Recognition}, pp.\  6538--6547, 2020.

\end{thebibliography}

\newpage
\appendix
\onecolumn

\section{Setup}

\subsection{Datasets}
\label{app:datasets}

\paragraph{MUG.}
Introduced by \citep{aifanti2010mug}, MUG encompasses a collection of image sequences featuring 52 subjects exhibiting six distinct facial expressions, namely, anger, fear, disgust, happiness, sadness, and surprise. Each video within the dataset is composed of a variable number of frames, ranging from 50 to 160. To standardize sequence length, as previously demonstrated by \citep{bai2021contrastively}, we employed a procedure in which 15 frames were randomly sampled from the original sequences. Subsequently, Haar Cascades face detection was implemented to isolate facial regions, resizing them to dimensions of $64 \times 64$ pixels. This process resulted in sequences denoted as $x \in \mathbb{R}^{15 \times 3 \times 64 \times 64}$. The post process dataset size is approximately $3500$ samples.

\paragraph{Sprites.}
Introduced by \citep{reed2015deep}. This dataset features animated cartoon characters with both static and dynamic attributes. The static attributes present variations in skin, tops, pants, and hair color, each offering six possible options. The dynamic attributes involve three distinct types of motion (walking, casting spells, and slashing) that can be executed in three different orientations (left, right, and forward). In total, the dataset comprises 1296 unique characters capable of performing nine distinct motions. Each sequence within the dataset comprises eight RGB images, each with dimensions of $64 \times 64$ pixels. We follow previous work protocol and we partitioned the dataset into 9000 samples for training and 2664 samples for testing.

\paragraph{PhysioNet.}
The PhysioNet ICU Dataset \citep{goldberger2000physiobank} is a medical time series dataset, encompassing the hospitalization records of 12,000 adult patients in the Intensive Care Unit (ICU). This comprehensive dataset incorporates time-dependent measurements, comprising physiological signals and laboratory data, alongside pertinent patient demographics, including age and the rationale behind their ICU admission. Additionally, the dataset is augmented with labels signifying in-hospital mortality events. Our pre-processing methodology aligns with the protocols outlined in \citep{tonekaboni2022decoupling}.

\paragraph{Air Quality.}
The UCI Beijing Multi-site Air Quality dataset, as detailed by \citep{zhang2017cautionary}, is a collection of hourly measurements of various air pollutants. These measurements were acquired over a four-year period, spanning from March 1st, 2013, to February 28th, 2017, from 12 nationally regulated monitoring sites. To complement this data, meteorological information for each site has been paired with the nearest weather station of the China Meteorological Administration. In alignment with the methodology outlined by~\citep{tonekaboni2022decoupling}, our experimental approach involves data pre-processing, entailing the segmentation of samples based on different monitoring stations and months of the year.

\paragraph{ETTh1.}
The ETTh1 is a subset of the Electricity Transformer Temperature (ETT) dataset, focusing on 1-hour-level data. It contains two years' worth of data from two Chinese counties. The goal is Long Sequence Time-Series Forecasting (LSTF) of oil temperature in transformers. Each data point includes the target value (oil temperature) and 6 power load features. The data is split into train, validation, and test sets, with a 12/4/4-month split ratio. 

\paragraph{TIMIT.}
Introduced by~\citep{garofolo1993timit} the dataset is a collection of read speech, primarily intended for acoustic-phonetic research and various speech-related tasks. This dataset encompasses a total of 6300 utterances, corresponding to approximately 5.4 hours of audio recordings. Each speaker contributes 10 sentences, and the dataset encompasses a diverse pool of 630 speakers, including both adult men and women. To facilitate data pre-processing, we adopt a methodology akin to previous research conducted by \citep{li2018disentangled}. Specifically, we employ spectrogram feature extraction with a 10ms frame shift applied to the audio. Subsequently, segments of 200ms duration, equivalent to 20 frames, are sampled from the audio and treated as independent samples.

\clearpage
\subsection{Metrics}
\label{app:metrics}
\paragraph{Accuracy}
This metric is the common evaluation protocol for assessing a model's capacity to preserve fixed features while generating others. Specifically, it entails the isolation of dynamic features while sampling static ones. This metric is evaluated by employing a pre-trained classifier, referred to as 'C' or the 'judge,' which has been trained on the same training dataset as the model. Subsequently, the classifier's performance is tested on the identical test dataset as the model. For example, in the case of the MUG dataset, the classifier examines the generated facial expression and verifies that it remains consistent during the sampling of static features. We refer this metric along the paper as "Acc" or "Accuracy Dynamic". In addition, we present in the ablation study the metric "Accuracy Static". This metric is exactly the same, just with one small modification. Fixing the static and sampling dynamic. 

\paragraph{Inception Score ($IS$).}
 This is a metric for the generator performance. First, we apply the judge on all the generated sequences $x_{1:T}$. Thus, getting $p(y | x_{1:T})$ which is the conditional predicted label distribution. Second, we take $p(y)$ which is the marginal predicted label distribution and we calculate the KL-divergence $\text{KL}[p(y | x_{1:T}) \, || \, p(y)]$. Finally, we compute $IS = \exp\left(\mathbb{E}_x \text{KL}[p(y | x_{1:T}) \, || \, p(y)]\right)$.

\paragraph{Inter-Entropy ($H(y|x)$).}
Inter-Entropy, often referred to as \(H(y|x)\), serves as a metric that reflects the confidence of a classifier (\(C\)) in making label predictions. A low value of Inter-Entropy indicates high confidence in the predictions made by the classifier. To measure this, we input \(k\) generated sequences into the classifier and compute the average entropy over these sequences, given by \(\frac{1}{k}\sum_{i=1}^{k}H(p(y | x_{i_{1:T}}))\).

\paragraph{Intra-Entroy ($H(y)$).}
Intra-Entropy, denoted as \(H(y)\), is a metric that measures the diversity among generated sequences. A high Intra-Entropy score indicates a high level of diversity among the generated data. This metric is computed by first taking a generated sample from the learned prior distribution \(p(y)\) and then applying a judge to obtain the predicted labels \(y\). The entropy of this label distribution \(H(y)\) quantifies the variability or uncertainty in the generated sequences.

\paragraph{AUPRC.}
The AUPRC (Area Under the Precision-Recall Curve) metric quantifies the precision-recall trade-off by measuring the area under the curve of the precision vs recall. A higher AUPRC indicates better model performance, with values closer to 1 being desirable, indicating high precision and high recall. 

\paragraph{AUROC.}

The AUROC (Area Under the Receiver Operating Characteristic Curve) metric quantifies the true positive (TPR) vs false positive (FRR) trade-off by measuring the area under the curve of the those rates. A higher AUPRC indicates better model performance, with values closer to 1 being desirable, indicating high precision and high recall.

\paragraph{MAE.}

The Mean Absolute Error (MAE) metric is a fundamental and widely-used measure in the field of regression analysis and predictive modeling. It quantifies the average magnitude of errors between predicted and actual values, providing a straightforward and intuitive assessment of model accuracy. Computed as the average absolute difference between predicted and true values, MAE is robust to outliers and provides a clear understanding of the model's precision in making predictions.

\paragraph{EER.}
The Equal Error Rate (EER) metric is a vital evaluation measure employed in the context of the speaker verification task, particularly when working with the Timit dataset. EER quantifies the point at which the false positive rate and false negative rate of a model in the speaker verification task are equal. It provides a valuable assessment of the model's performance, specifically in the context of speaker recognition. 

\clearpage
\subsection{Hyper-parameters}
\label{app:hyperparameters}

We compute the following objective function:

\begin{equation}  
\mathcal{L} = \max_{\theta,\phi,\psi} \; \mathbb E_{p_D} (\mathcal{L}_\text{recon} - \mathcal{L}_\text{reg}) \ .
\end{equation}

\begin{equation} 
    \mathcal{L}_\text{reg} = \beta \; \kldiv{\dist{q}{s}{x_1 \, ; \, \phi_s}}{p(s)} + \beta \; \kldiv{\dist{q}{d_{2:T}}{x_{2:T} \, ; \, \phi_d}}{p(d_{2:T} \, ; \, \psi)} \ ,
\end{equation}

\begin{equation}
    \mathcal{L}_\text{recon} = \mathbb{E}_{s \sim q_{\phi_s}} [ \mathbb{E}_{d_{2:T} \sim q_{\phi_d}} \log \dist{p}{x_{2:T}}{s, d_{2:T} \, ; \, \theta}] 
    + \alpha \, \mathbb{E}_{s \sim q_{\phi_s}} [\log \dist{p}{x_1}{s, d_1 \, ; \, \theta} ] \ ,
\end{equation}


We determined optimal hyper-parameters, specifically $\alpha$ for the reconstruction loss and $\beta$ for the static KL term, using HyperOpt to search for values in the range of Zero to One. It's important to note that we did not normalize the mean squared error (MSE) loss by the batch size during this process. Additionally, optimization was carried out using the Adam optimizer with a learning rate chosen from $0.001, 0.0015, 0.002, 0.003$ and we considered feature dimensions of $8, 16, 32$ for time series datasets and $64, 110, 256$ for the image and audio datasets for both static and dynamic dimensions. A comprehensive summary of these optimal hyper-parameters for each task and dataset is available in Tab.~\ref{tab:dataset_hyperparameters}, and all training processes were limited to a maximum of 2000 epochs.

\begin{table*}[ht]
\centering
\caption{Dataset Hyper-parameters.}
\vskip 0.1in
\label{tab:dataset_hyperparameters}
\footnotesize
\begin{tabular}{ccccccccc}
\toprule
\textbf{Dataset} & $\alpha$ & $\beta$ & \textbf{Learning Rate} & \textbf{Batch Size} & \textbf{Static ($s_d$)} & \textbf{Dynamic ($d_d$)} \\
\midrule
MUG & 0.35 &  0.28  & $1.5\times10^{-3}$ & 64 & 64 & 64 \\
PhysioNet & 0.1 & 0.01 &$3\times10^{-3}$ & 30 & 8 & 8 \\
Air Quality & 0.28 & $2\times10^{-4}$ & $3\times10^{-3}$ & 10 & 8 & 8 \\
ETTh1 & 0.23 & 0.03 & $1\times10^{-3}$ & 10 & 8 & 8 \\
Timit & $6\times10^{-4}$ & $6\times10^{-4}$ & $1\times10^{-3}$ & 10 & 110 & 110 \\
Sprites & 0.2 & 0.2 & $2\times10^{-3}$ & 128 & 256 & 256 \\

\bottomrule
\end{tabular}
\vskip -0.1in
\end{table*}

\subsection{Architecture}
\label{app:architecture}

In Fig. \ref{fig:arch} we present our method architecture. Generally, our architecture comprises of three components. The encoder, the disentanglement module and the decoder. The disentanglement module is similar to every data modality and it is fully explained in the figure in the main text. The encoder and the decoder for each data modality are different, and in what follows, we present the architecture details for each one of them.  


\paragraph{Video:}  
\begin{enumerate}
    \item \textit{Encoder} - The encoder comprises of 5 layers of Conv2d followed by BatchNorm2D followed by LeakyReLU. Below are the Conv2D (input channel dimension, output channel dimension, kernel size, stride,  padding) hyper-parameters given for a $64 \times 64 \times 3$ input image, ordered as they appear in the encoder: $(3, 32, 4, 2, 1)\rightarrow(32, 64, 4, 2, 1) \rightarrow (64, 128, 4, 2, 1) \rightarrow  (128, 256, 4, 2, 1) \rightarrow (256, 128, 4, 2, 1)$. 

    \item \textit{Decoder} - Similarly to the encoder, the decoder comprises of 5 layers. The first 4 layers are Conv2DTranspose followed by BatchNornm2D followed by LeakyRelu. The final layer is Conv2D followed by BatchNorm2D followed by a Sigmoid function. Below are the Conv2DTranspose (input channel dimension, output channel dimension, kernel size, stride,  padding) hyper-parameters given for a latent code with size of the concatenated static and dynamic $s_d + d_d$ by the order of their appearance in the decoder: $(s_d + d_d, 256, 4, 1, 0)\rightarrow(256, 128, 4, 1, 0) \rightarrow (128, 64, 4, 1, 0) \rightarrow  (64, 32, 4, 1, 0) \rightarrow (32, 3, 4, 1, 0)$. 
\end{enumerate}

\paragraph{Time Series:} When training for this data, we adopt a strategy that involves selecting a window of data points, rather than a single data point for the static part. This approach is tailored to each dataset, with the window size varying based on the dataset's length. Specifically, for the Air Quality and ETTh1 datasets, we utilize 28 distinct windows, each encompassing 24 data points. For the Physionet dataset, our methodology involves the use of 20 windows, where each window comprises of 4 data points. We found that this window-based approach enables a more comprehensive analysis of time series data.

\begin{enumerate}
    \item \textit{Encoder} - The encoder consists of three linear layers with the next dimensions: $(10, 32) \rightarrow (32, 64) \rightarrow (64, 32)$. ReLU activations are applied after each linear layer.
    
    \item  \textit{Decoder} - The decoder is a linear layer that projects the latent codes into a 32-dimensional space, followed by a tanh activation function. Subsequently, for the different tasks, the output is passed through an LSTM with a hidden size of 32. The LSTM's output is fed into two linear layers, each followed by a ReLU activation: \texttt{Linear(32, 64)} and \texttt{Linear(64, 32)}. Finally, the output is projected through two more linear layers to produce the mean and covariance parameters, which are used to sample the final output. 
\end{enumerate}

\section{Additional Experiments and details} 

\begin{figure*}[t]
    \centering
    \includegraphics[width=1\linewidth]{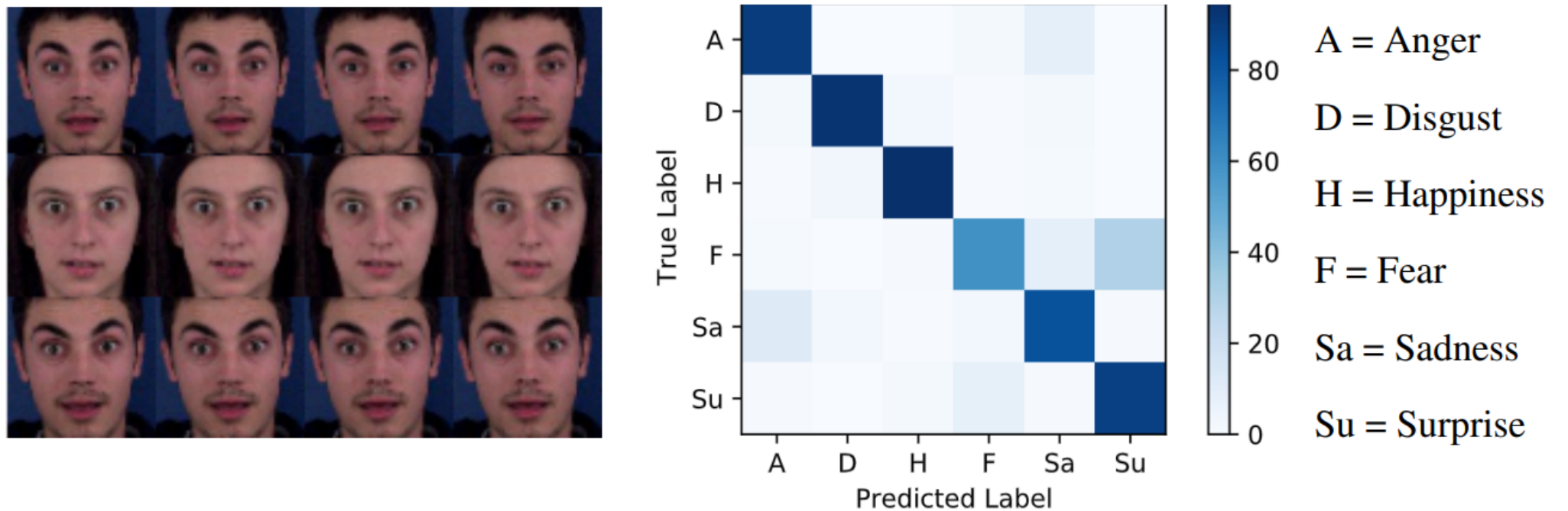}
    \vspace{-5mm}
    \caption{Different emotions may manifest similarly for different people, as seen in the fear expression of one person (left, top row) and surprise expression of another person (left, middle row). We
further quantify this effect in a confusion matrix (right).}
    \label{fig:mug_failure_cases}
\end{figure*}

\subsection{Failure case analysis on MUG.}
\label{app:failure_case_mug}
The MUG dataset is a common disentanglement benchmark, comprising of six expressions made by $52$ different subjects. Tab.~\ref{tab:gen_sprites_mug} shows that all models fail to produce $>90\%$ accuracy for this dataset. In what follows, we explore the failure cases of our model, toward characterizing where and why the model fails. To this end, we observe that \emph{different} facial expressions may look \emph{similar} for different subjects or even for the same subject. For example, we consider two persons, $h_1$ and $h_2$ and we plot their facial expressions in Fig.~\ref{fig:mug_failure_cases} (left). The top row is the surprised expression of $h_1$, the middle row is the fear expression of $h_2$, and the bottom row is the fear expression of $h_1$. Clearly, all rows are similar to the human eye, and thus, unsupervised methods such as the one we propose will naturally be challenged by such examples. To quantify this effect we compute the confusion matrix of our model benchmark predictions in Fig.~\ref{fig:mug_failure_cases} (right). For each cell, we estimate the ratio of predicted labels (columns) vs. the true labels (rows). Thus, the main diagonal represents the true positive rate, whereas off diagonal cells encode false positive and false negative cases. Ideally, we would like to have $100\%$ (dark blue) on the main diagonal, which means that our disentangled generations are perfect. While this is largely the case for Anger, Disgust and Happiness, we find that Fear, Sadness and Surprise may be confounded. For instance, there is $\approx 35\%$ confusion between Fear and Surprise. Our analysis suggests that hierarchical disentanglement based first on dominant features such as subject identity, and only then on facial expression may be a strong inductive bias in modeling disentanglement~\citep{scholkopf2021toward}. We leave further consideration and exploration of this aspect to future work.

\subsection{Audio Dataset - TIMIT}

\paragraph{Experiment description.} 
\label{app:timit_experiment}
To show the robustness of sequential disentanglement methods, previous works have used a common audio verification benchmark on the TIMIT dataset. We evaluate our model using the common benchmark protocol described in \citep{li2018disentangled} and with the same encoding and decoding architecture for a fair comparison. Briefly, the evaluation protocol is given two audio tracks, the goal of this task is to recognize if they come from the same speaker. Unrelated to what the words context are. The verification is done by EER - Equal Error Rate \citep{chenafa2008biometric} metric where we use cosine similarity and $\epsilon \in [0, 1]$ threshold. Given two audio tracks $x^1_{1:T} , x^2_{1:T}$ we extract their static $s^1, s^2$ representations. We follow the exact extraction process as in \citep{bai2021contrastively, li2018disentangled}. Then, we measure the cosine similarity of the pairs, if it is higher than $\epsilon$, we classify them as the same speaker, and as different speakers otherwise. The $\epsilon$ is determined by calibration of the EER. We conducted the above procedure once for the static features and separately for the dynamic features, then we report each experiment's EER. In the static setup, lower error is desired, since it should encapsulate information about the speaker. On the other hand, in the dynamic setup the higher the error the better since it should not encapsulate information about the speaker. In addition, we show the information leakage gap, the gap between the static and dynamic setups, which indicates the quality of the disentanglement. We report our model performance in Tab.~\ref{tab:timit}. Notably, our model achieves state-of-the-art disentanglement gap, surpassing by approximately $1.3 \%$ the best previous method. 

\begin{table}[t]
    \centering
    \caption{TIMIT voice verification task.}
    \vskip 0.1in
    \label{tab:timit}
    \begin{tabular}[t]{lccc}
        \toprule
        Method & static EER$\downarrow$ & dynamic EER $\uparrow$ & Disentanglement Gap $\uparrow$ \\
        \midrule
        FHVAE & $5.06\%$ & $22.77\%$ & $17.71\%$ \\
        DSVAE & $5.64\%$ & $19.20\%$ & $13.56\%$ \\
        R-WAE & $4.73\%$ & $23.41\%$ & $18.68\%$ \\
        S3VAE & $5.02\%$ & $25.51\%$ & $20.49\%$ \\
        SKD   & $4.46\%$ & $26.78\%$ & $22.32\%$ \\
        C-DSVAE & $4.03\%$ & $31.81\%$ & $27.78\%$ \\
        SPYL & $\boldsymbol{3.41\%}$ & $33.22 \%$ & $29.81 \%$  \\
        \midrule
        Ours & $3.50\%$ & $\boldsymbol{34.62 \%}$ & $\boldsymbol{31.11 \%}$ \\
        \bottomrule
    \end{tabular}
    \vskip -0.1in
\end{table}

\subsection{Ablation Studies}
\label{app:ablation_t_cont}
Due to space limitations, we extend here the ablation study results that were reported in the main text at Sec.~\ref{subsec:ablt_studies}.

\paragraph{Dependence on the $i$-th sample.} In Tab.~\ref{tab:ablation_ts_data_app_t} and Tab.~\ref{tab:ablation_frame_mug}, we present the results of our ablation study on frame selection across multiple datasets on classification tasks. The comparison is made between the first frame, the middle frame, and the last frame. Notably, the outcomes indicate a high degree of consistency among these frames, showcasing the robustness of our frame selection approach. The similarity in results across different frames suggests that our model is not overly sensitive to the specific choice of frames.


\begin{table}[t]
    \centering
    \caption{Additional ablation study results on the time series benchmark.}
    \vskip 0.1in
    \begin{tabular}{l @{\hskip 0.5cm} c @{\hskip 0.5cm} c}
        \toprule
        \textbf{Method} & \textbf{Accuracy (PhysioNet)} $\uparrow$ & \textbf{Accuracy (Air Quality)} $\uparrow$ \\
        \midrule
        $x_{rob}$  & $40.81 \% \pm 0.328$ & $43.73 \% \pm 0.15$ \\
        $x_{random}$  & $54.87 \% \pm 0.138$ & $62.73 \% \pm 0.1$ \\
        $x_{T/2}$  & $55.20 \% \pm 0.120$ & $66.73 \% \pm 0.07$ \\
        $x_{T}$   & $55.19 \% \pm 0.260$ & $69.53 \% \pm 0.22$  \\
        \midrule
        $x_{1}$ & $56.87 \% \pm 0.34$ & $65.87 \% \pm 0.01$ \\
        \bottomrule
    \label{tab:ablation_ts_data_app_t}
    \end{tabular}
    \vskip -0.1in
\end{table}

\begin{table}[t]
    \centering
    \caption{Additional ablation study results on the MUG benchmark.}
    \vskip 0.1in
    \begin{tabular}{l @{\hskip 0.5cm} c @{\hskip 0.5cm} c  @{\hskip 0.5cm} c  @{\hskip 0.5cm} c}
        \toprule
        & \multicolumn{2}{c}{Static Features} & \multicolumn{2}{c}{Dynamic Features} \\
        \textbf{Method} & \textbf{Static Acc1} $\uparrow$ & 
        \textbf{Dynamic Acc1} $\downarrow$ &
        \textbf{Static Acc2} $\downarrow$ &
        \textbf{Dynamic Acc2} $\uparrow$ \\
        \midrule
        $x_{rob}$  & $99.53 \%$ & $29.45 \%$  & $3.35 \%$  & $78.39 \%$ \\
        $x_{random}$  & $99.53 \%$ & $22.91 \%$  & $2.98 \%$  & $85.55 \%$ \\
        $x_{T/2}$  & $99.71 \%$ & $22.32 \%$  & $3.01 \%$  & $85.42 \%$ \\
        $x_{T}$   & $99.67 \%$ & $22.42 \%$  & $3.06 \%$ & $85.35 \%$ \\
        \midrule
        $x_{1}$ & $99.42 \%$ & $20.85 \%$  & $2.89 \%$ & $86.90 \% $\\
        \bottomrule
    \label{tab:ablation_frame_mug}
    \end{tabular}
    \vskip -0.1in
\end{table}

\begin{figure*}[ht]
    \centering
    \includegraphics[width=1\linewidth]{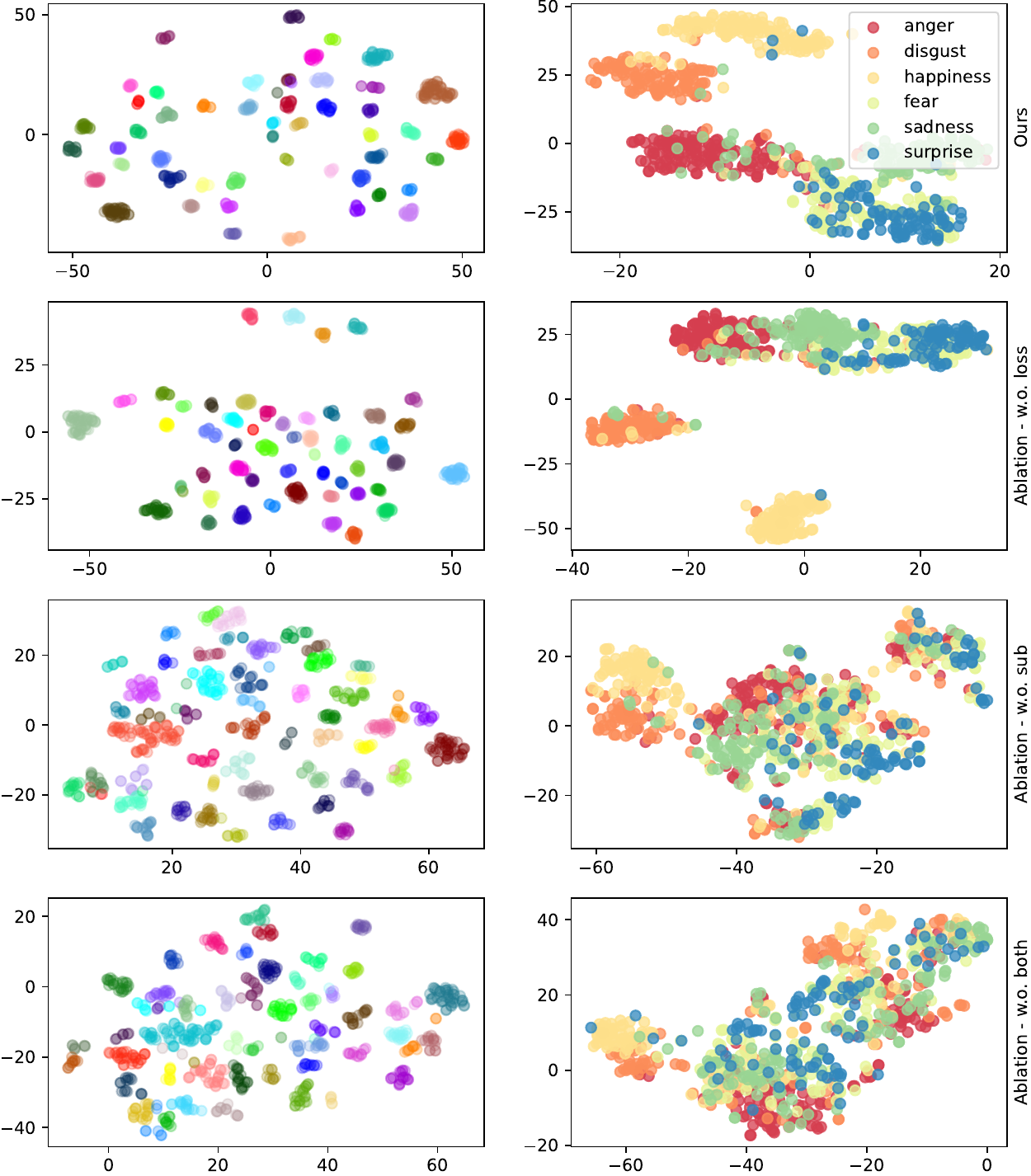}
    \caption{t-SNE plots on MUG dataset of the latent static and dynamic factors. Latent static codes, colored by subject identity (left), and latent dynamic codes, colored by dynamic attribute (right).} 
    \label{fig:tsne_mug_ablation}
\end{figure*}

\paragraph{Static and dynamic clustering.} In this section, we expand our qualitative assessment of both the static $s$ and dynamic $d_{1:T}$ embeddings within the latent space of our approach. Specifically, we aim to demonstrate the significance of our loss and subtraction choices on the overall performance of our model. In Fig.~\ref{fig:tsne_mug_ablation}, we use t-SNE to project the static and dynamic embeddings into a two-dimensional space. The static embeddings are depicted in orange, while the dynamic embeddings are shown in blue (left). Additionally, we independently project each of the factors. This process is repeated for each of the ablated models, wherein we eliminate the loss function, the subtraction module, and both of them. It becomes apparent that in the absence of the loss function, the 'anger' dynamics overlap with 'sadness'; similarly, 'fear' and 'surprise' overlap, consistent with our previous observations. Moreover, when we remove the subtraction module (third row) or eliminate both components (last row), a notable decline in performance is observed. In these scenarios, the dynamic embeddings become cluttered, with different dynamics overlapping, indicating poor clustering and disentanglement. In conclusion, in addition to the notable decline in performance we observe in the quantitative evaluation, we demonstrate how important the additional loss term and subtraction are through qualitative assessment.

\subsection{Disentanglement Generation with Standard Deviation}
\label{generative_video_std}
For simplicity, we present in Tab.~\ref{tab:gen_sprites_mug} in the main text the results without standard deviation. Due to the nature of generative models to produce unstable results it is important to validate that a model is stable and shows a statistically significant improvement. Therefore, we present full results of our model with standard deviation measures in Tab.~\ref{tab:gen_sprites_mug_std}. We repeat the task 300 times with different seeds and report its mean and standard deviation. The results show that our model is profoundly stable.

\begin{table*}[!t]
    \centering
    \caption{Sprites and MUG datasets with standard deviation measures.}
    \vskip 0.1in
    \label{tab:gen_sprites_mug_std}
    \footnotesize
    \begin{tabular}[t]{l|ccccccc|c}
        \toprule
         & MoCoGAN & DSVAE & R-WAE & S3VAE & SKD & C-DSVAE & SPYL & Ours \\
        \midrule
        \multicolumn{1}{l|}{\textbf{Sprites}} & & & & & & & & \\
        Acc$\uparrow$ & $92.89\%$ & $90.73\%$ & $98.98\%$ & $99.49\%$ & $\boldsymbol{100\%}$ & $99.99\%$ & $\boldsymbol{100\% \pm 0}$ & $\boldsymbol{100\% \pm 0}$ \\
        IS$\uparrow$ & $8.461$ & $8.384$ & $8.516$ & $8.637$ & $\boldsymbol{8.999}$ & $8.871$ & $8.942 \pm 3.3 \mathrm{e}{-5}$ & $8.942 \pm 7 \mathrm{e}{-5}$ \\
        $H(y|x){\downarrow}$ & $0.090$ & $0.072$ & $0.055$ & $0.041$ & $\boldsymbol{\expnumber{1.6}{-7}}$ & $0.014$ & $0.006 \pm 4 \mathrm{e}{-6}$ & $0.006 \pm 3 \mathrm{e}{-6}$ \\
        $H(y){\uparrow}$ & $2.192$ & $2.192$ & $\boldsymbol{2.197}$ & $\boldsymbol{2.197}$ & $\boldsymbol{2.197}$ & $\boldsymbol{2.197}$ & $\boldsymbol{2.197 \pm 0}$ & $\boldsymbol{2.197 \pm 0}$ \\
        \midrule
        \multicolumn{1}{l|}{\textbf{MUG}} & & & & & & & & \\
        Acc$\uparrow$ & $63.12\%$ & $54.29\%$ & $71.25\%$ & $70.51\%$ & $77.45\%$ & $81.16\%$ & $85.71\% \pm 0.9$ & $\boldsymbol{86.90\% \pm 0.9}$ \\
        IS$\uparrow$ & $4.332$ & $3.608$ & $5.149$ & $5.136$ & $5.569$ & $5.341$ & $5.548 \pm 0.039$ & $\boldsymbol{5.598 \pm 0.068}$ \\
        $H(y|x){\downarrow}$ & $0.183$ & $0.374$ & $0.131$ & $0.135$ & $0.052$ & $0.092$ & $0.066 \pm 4 \mathrm{e}{-3}$ & $\boldsymbol{0.041 \pm 8 \mathrm{e}{-3}}$ \\
        $H(y){\uparrow}$ & $1.721$ & $1.657$ & $1.771$ & $1.760$ & $1.769$ & $1.775$ & $1.779 \pm 6 \mathrm{e}{-3}$ & $\boldsymbol{1.782 \pm 0.013}$ \\
        \bottomrule
    \end{tabular}
    \vskip -0.1in
\end{table*}

\subsection{Information leakage gap in video with standard deviation measures}
Similar to App.\ref{generative_video_std}, we display the complete results of Tab.\ref{tab:full_dis_met_cls_task}. We repeat the same experiment protocol as in \cite{naiman2023sample}. The results are reported in Tab.\ref{tab:std_full_dis_met_cls_task}. The experiment results show the statistical significance of our model.

    \begin{table*}
        \caption{MUG dataset with standard deviation measures.}
        \vskip 0.1in
        \label{tab:std_full_dis_met_cls_task}
        \resizebox{1\textwidth}{!}{
        \centering
        \begin{tabular}[t]{ll|ccc|ccc}
            \toprule
            & & \multicolumn{3}{c|}{Static Features} & \multicolumn{3}{c}{Dynamic Features} \\
            Classifier & Method  & Static Acc $\uparrow$ & Dynamic Acc $\downarrow$ & Leakage Gap $\uparrow$ & Static Acc  $\downarrow$ & Dynamic Acc $\uparrow$ & Leakage Gap $\uparrow$ \\
            \midrule
            \multirow{4}{*}{Latent} & random & $1.92 \%$ & $16.66\%$ & - & $1.92 \%$ & $16.66\%$ & - \\
            & C-DSVAE    & $98.75\% \pm 1.1$ & $76.25\% \pm 3.6 $ & $22.25\%$ & $26.25\% \pm 3.4$ & $82.50\% \pm 2.1$ & $56.25\%$ \\
            & SPYL       & $98.12\% \pm 0.9 $ & $68.75\% \pm 4.5$ & $29.37\%$ & $\boldsymbol{10.00\% \pm 2.8}$ & $\boldsymbol{85.62\% \pm 1.5}$ & $\boldsymbol{\textcolor{blue}{75.62\%}}$ \\
            & Ours        & $\boldsymbol{99.35\% \pm 1.2}$ & $\boldsymbol{45.06\% \pm 3.8}$ & $\boldsymbol{\textcolor{blue}{54.29\%}}$ & $11.36\% \pm 0.6$ & $85.51\% \pm 2.3$ & $74.15\%$ \\
            \bottomrule
        \end{tabular}
        }
        \vskip -0.1in
    \end{table*}

\subsection{Generative Sampling}
\begin{figure}[ht]
    \centering
    \includegraphics[width=1\linewidth]{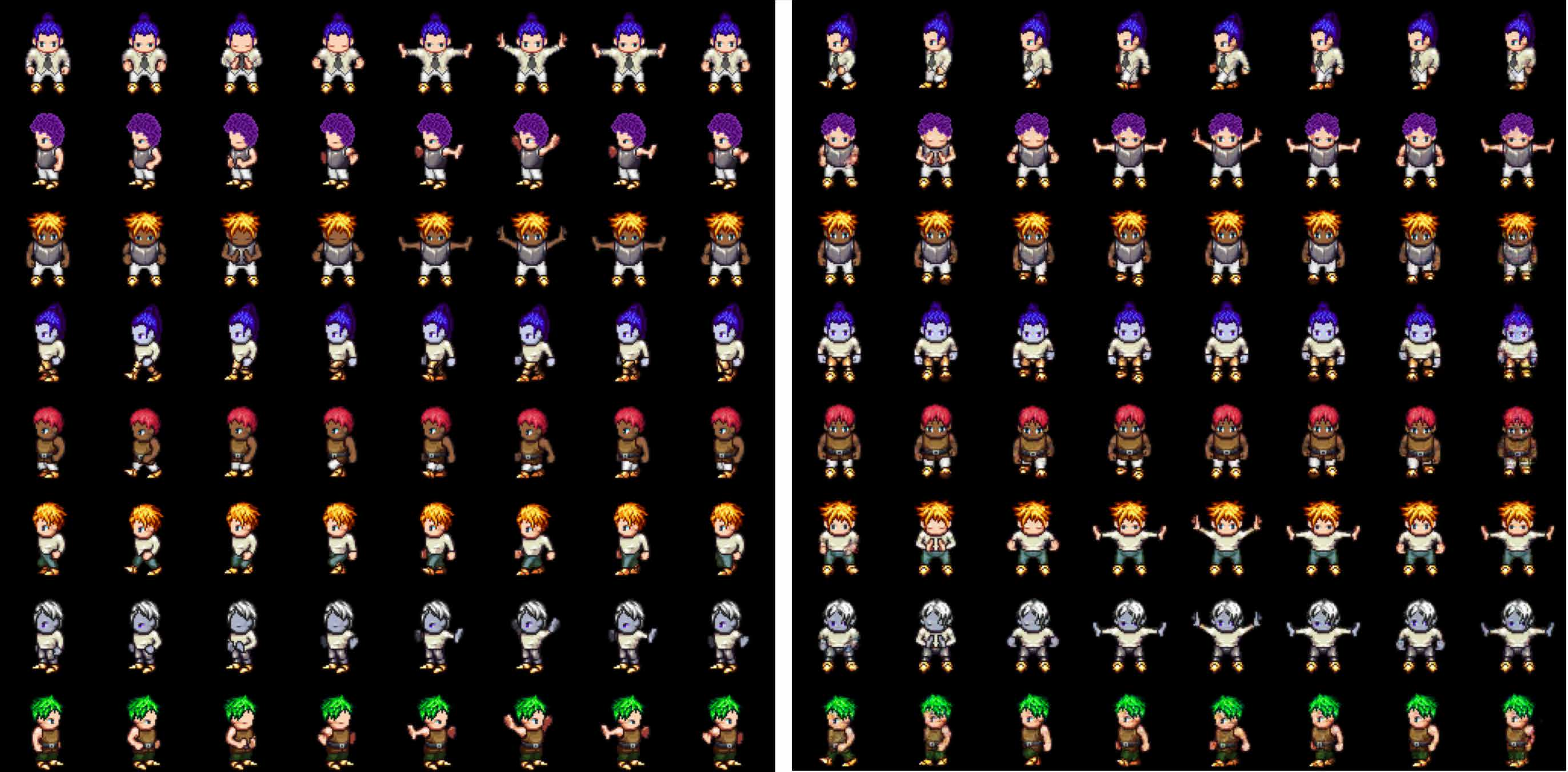}
    \caption{Dynamics generation results in Sprites dataset. On the left side, the original sequence. On the right side, the same sequence with different dynamics.} 
    \label{fig:sprites_generated_dynamic}
\end{figure}

\begin{figure}[ht]
    \centering
    \includegraphics[width=1\linewidth]{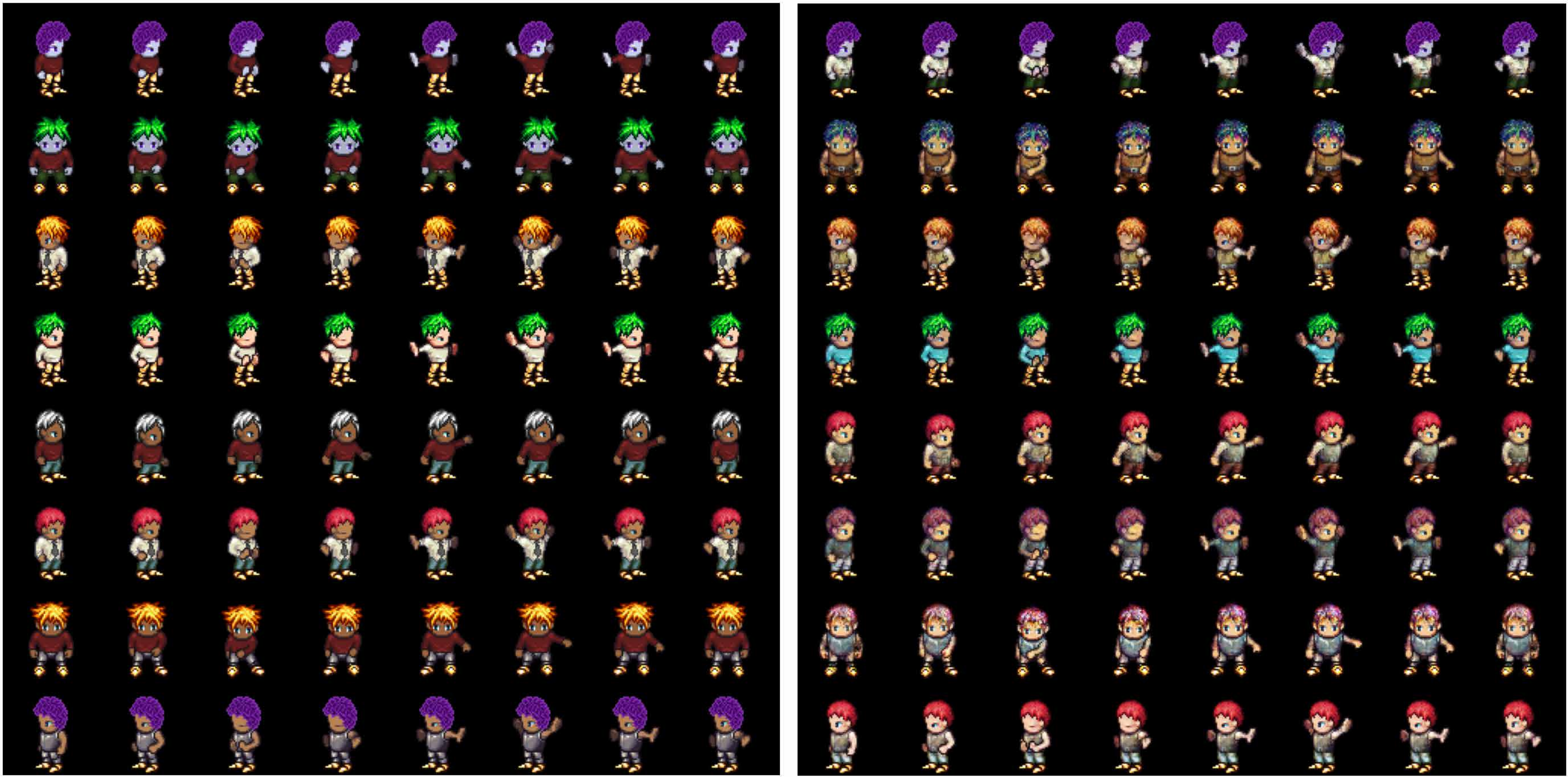}
    \caption{Static generation results in Sprites dataset. On the left side, the original sequence. On the right side, the same sequence with different static features.} 
    \label{fig:sprites_generated_static}
\end{figure}

\begin{figure}[ht]
    \centering
    \includegraphics[width=1\linewidth]{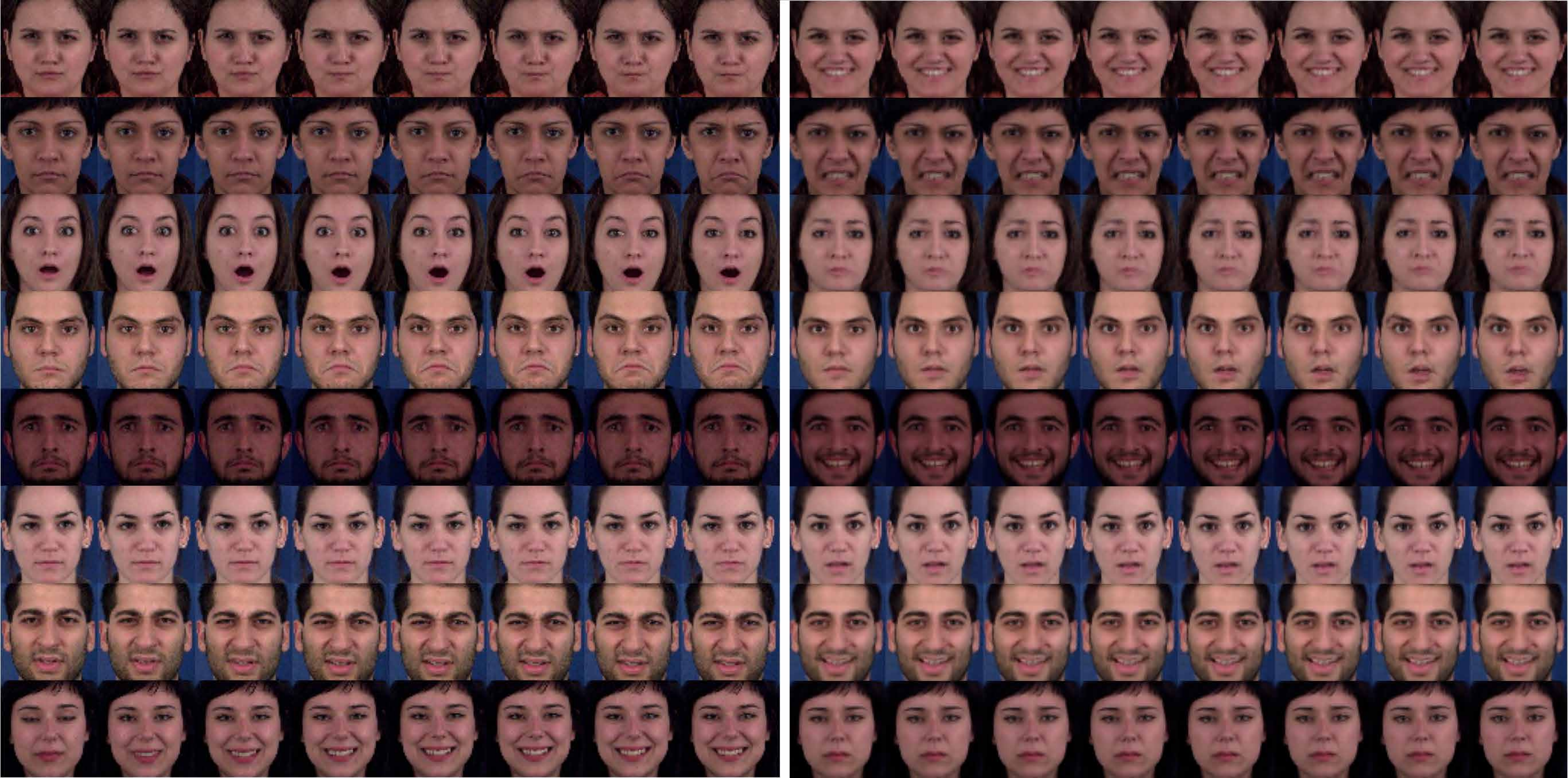}
    \caption{Dynamics generation results in MUG dataset. On the left side, the original sequence. On the right side, the same sequence with different dynamics.} 
    \label{fig:mug_generated_dynamic}
\end{figure}

\begin{figure}[ht]
    \centering
    \includegraphics[width=1\linewidth]{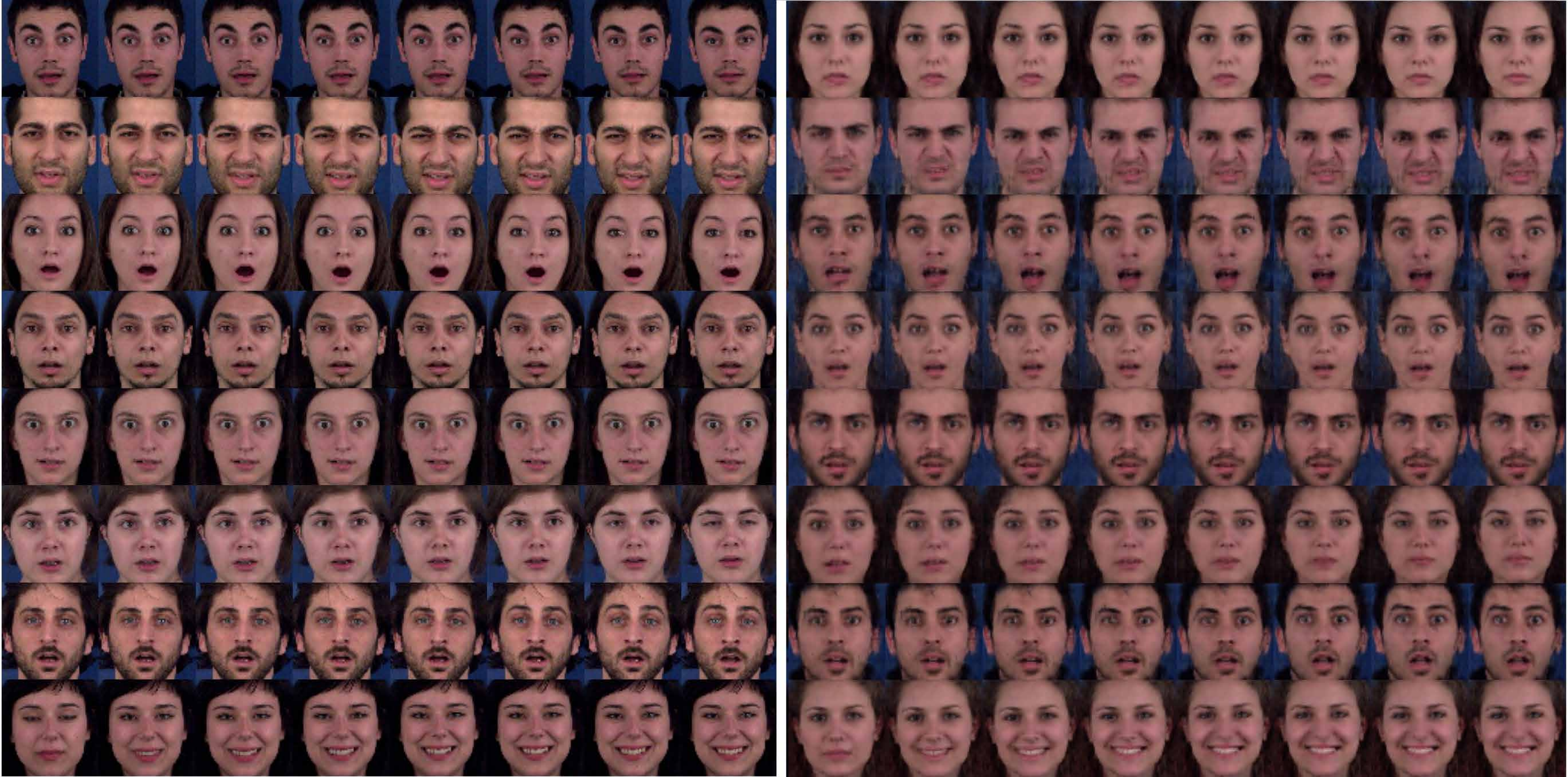}
    \caption{Static generation results in MUG dataset. On the left side, the original sequence. On the right side, the same sequence with different static features.} 
    \label{fig:mug_generated_static}
\end{figure}

In this experiment, we show qualitatively our model's capability to fix one factor and sample the other on the MUG and Sprites datasets. In Fig.~\ref{fig:sprites_generated_dynamic} and Fig.~\ref{fig:mug_generated_dynamic}, we fix the static component and sample new dynamic component. In Fig.~\ref{fig:sprites_generated_static} and Fig.~\ref{fig:mug_generated_static}, we fix the dynamic component and sample a new static component.

\subsection{Swap Examples}
\begin{figure}[ht]
    \centering
    \includegraphics[width=1\linewidth]{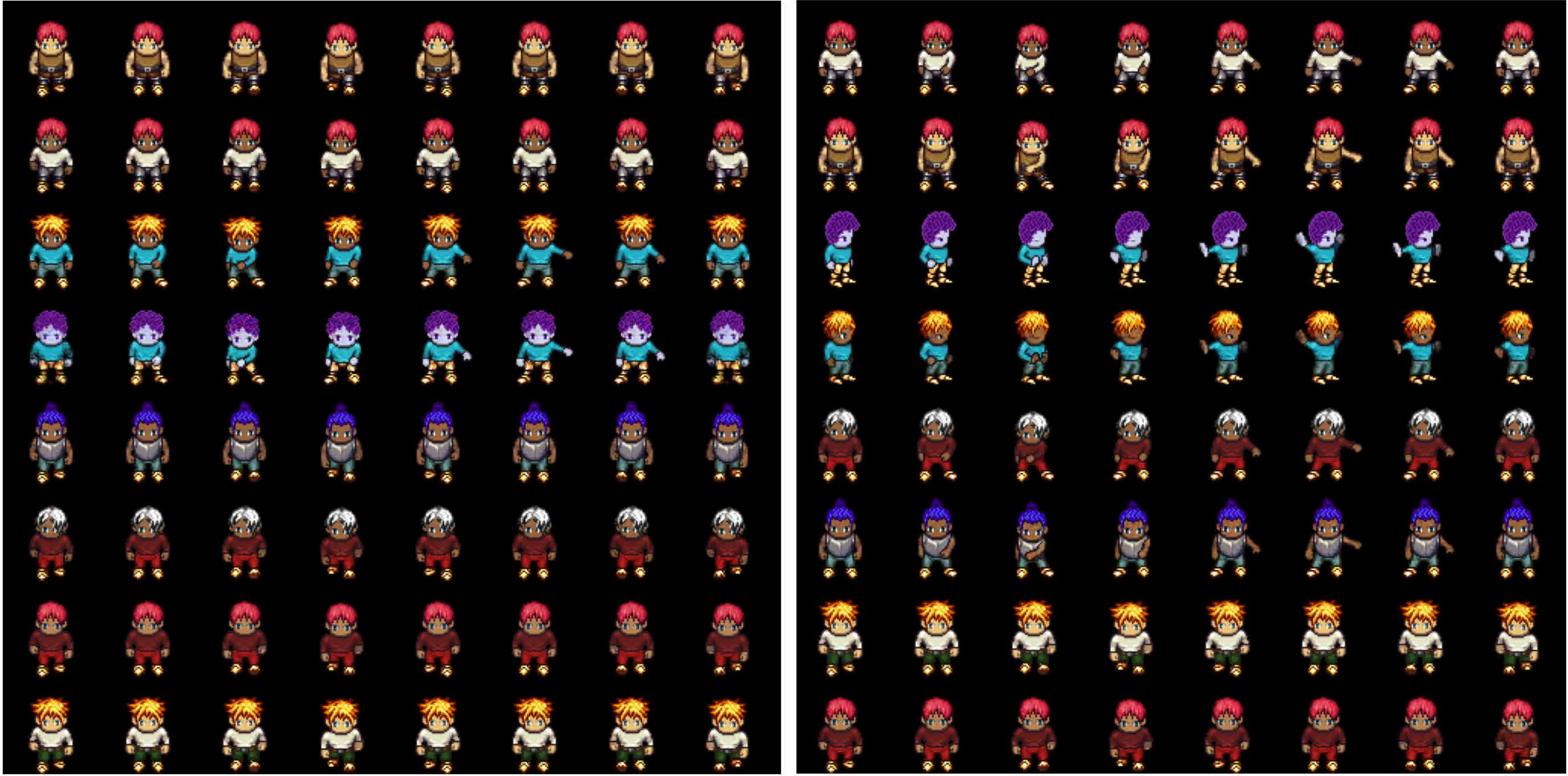}
    \caption{Swapping results in Sprites dataset. Each odd row (counting from One) contains Two original samples. Each even row contains the Two swapped samples of their above row.} 
    \label{fig:sprites_swap_content}
\end{figure}

\begin{figure}[ht]
    \centering
    \includegraphics[width=1\linewidth]{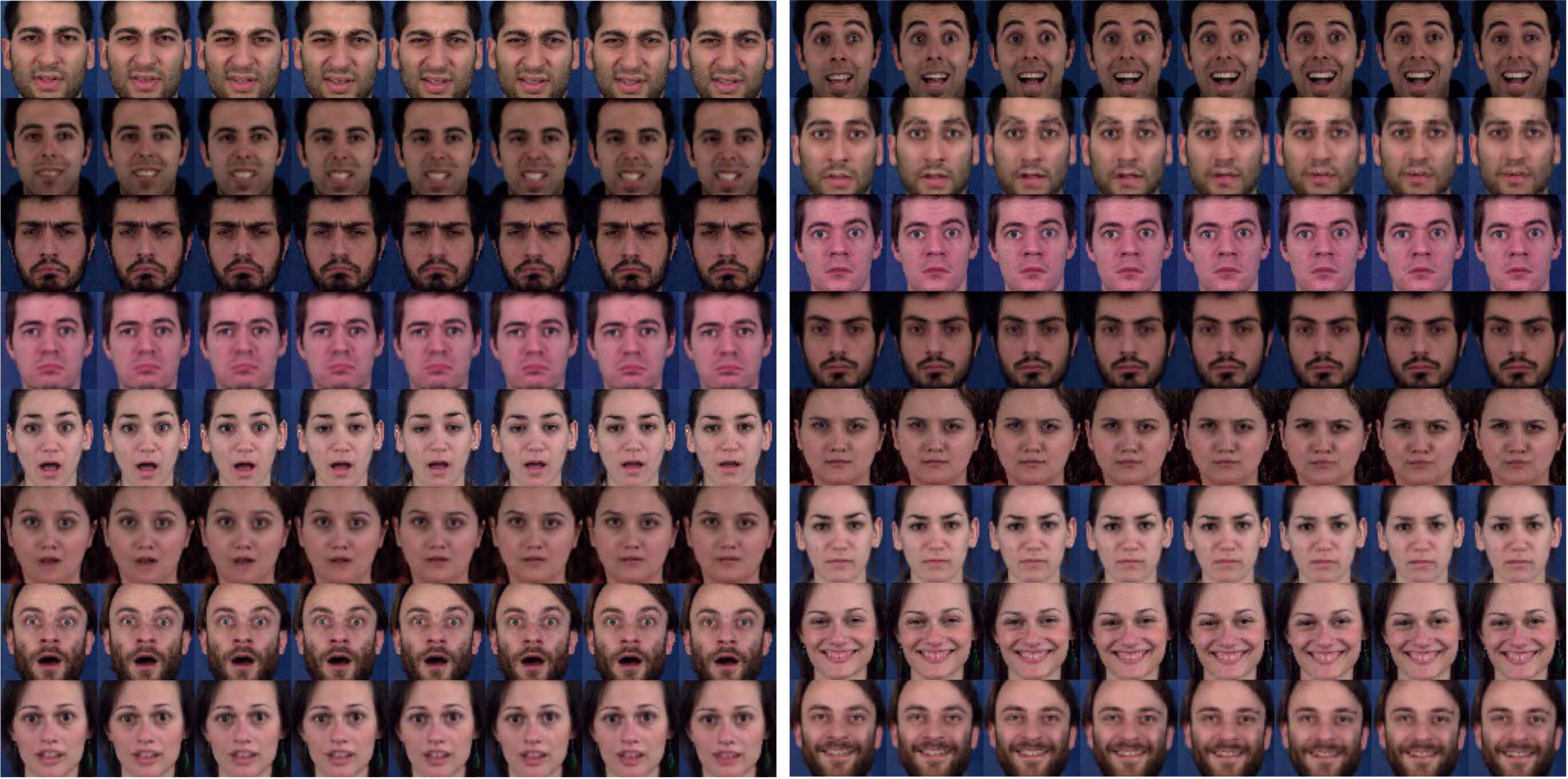}
    \caption{Swapping results in MUG dataset. Each odd row (counting from One) contains Two original samples. Each even row contains the Two swapped samples of their above row.} 
    \label{fig:swap_mug_1}
\end{figure}

We extend the swap experiment in the main text in Sec.~\ref{subsec:qualitative_eval} showing more swap examples between two samples on the MUG and the Sprites in Fig.~\ref{fig:sprites_swap_content} and Fig.~\ref{fig:swap_mug_1}.

\begin{figure}[h]
    \centering
    \includegraphics[width=1\linewidth]{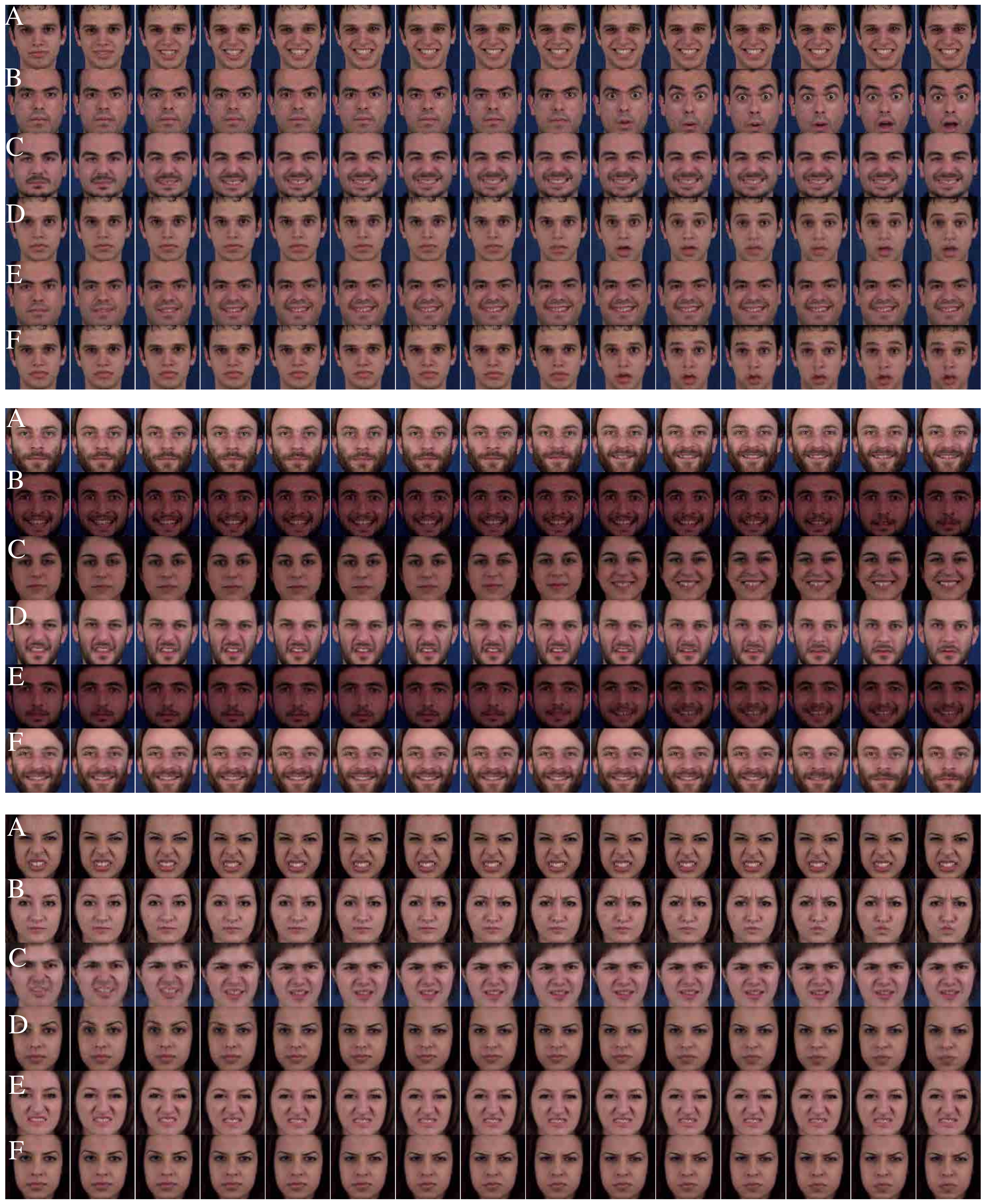}
    \caption{Qualitative example of swap between source and target sequences. A is the source, B is the target, C (E) is when static is swapped from source to target, and D (F) is when dynamics are swapped. C and D are the swaps for SPYL method and E and F are swaps of our method.} 
    \label{fig:swap_comp_1}
\end{figure}

\begin{figure}[h]
    \centering
    \includegraphics[width=1\linewidth]{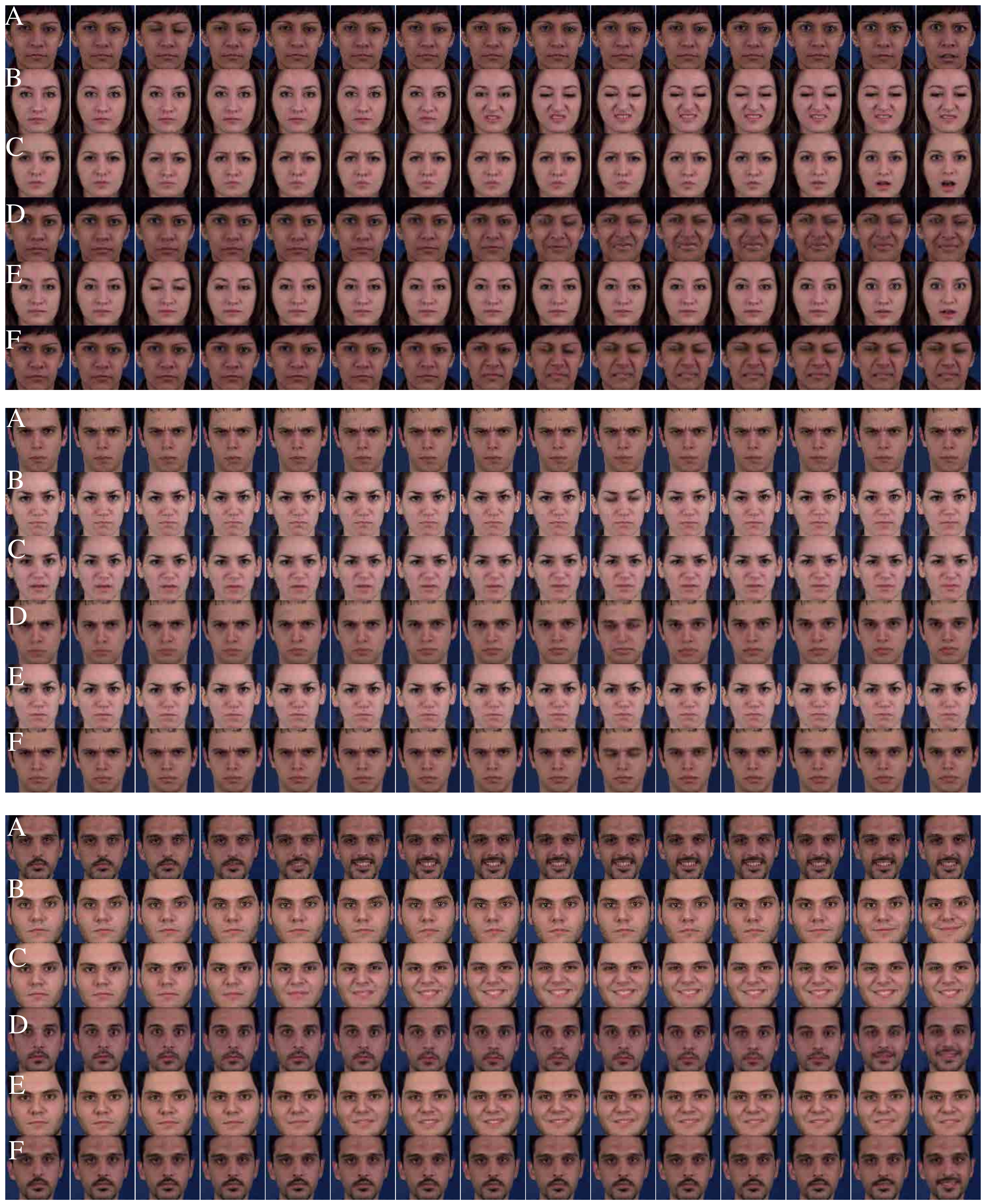}
    \caption{Qualitative example of swap between source and target sequences. A is the source, B is the target, C (E) is when static is swapped from source to target, and D (F) is when dynamics are swapped. C and D are the swaps for SPYL method and E and F are swaps of our method.} 
    \label{fig:swap_comp_2}
\end{figure}

\subsection{Time Series Reconstruction}
\begin{figure}[htbp]
    \centering
    \includegraphics[width=0.7\textwidth]{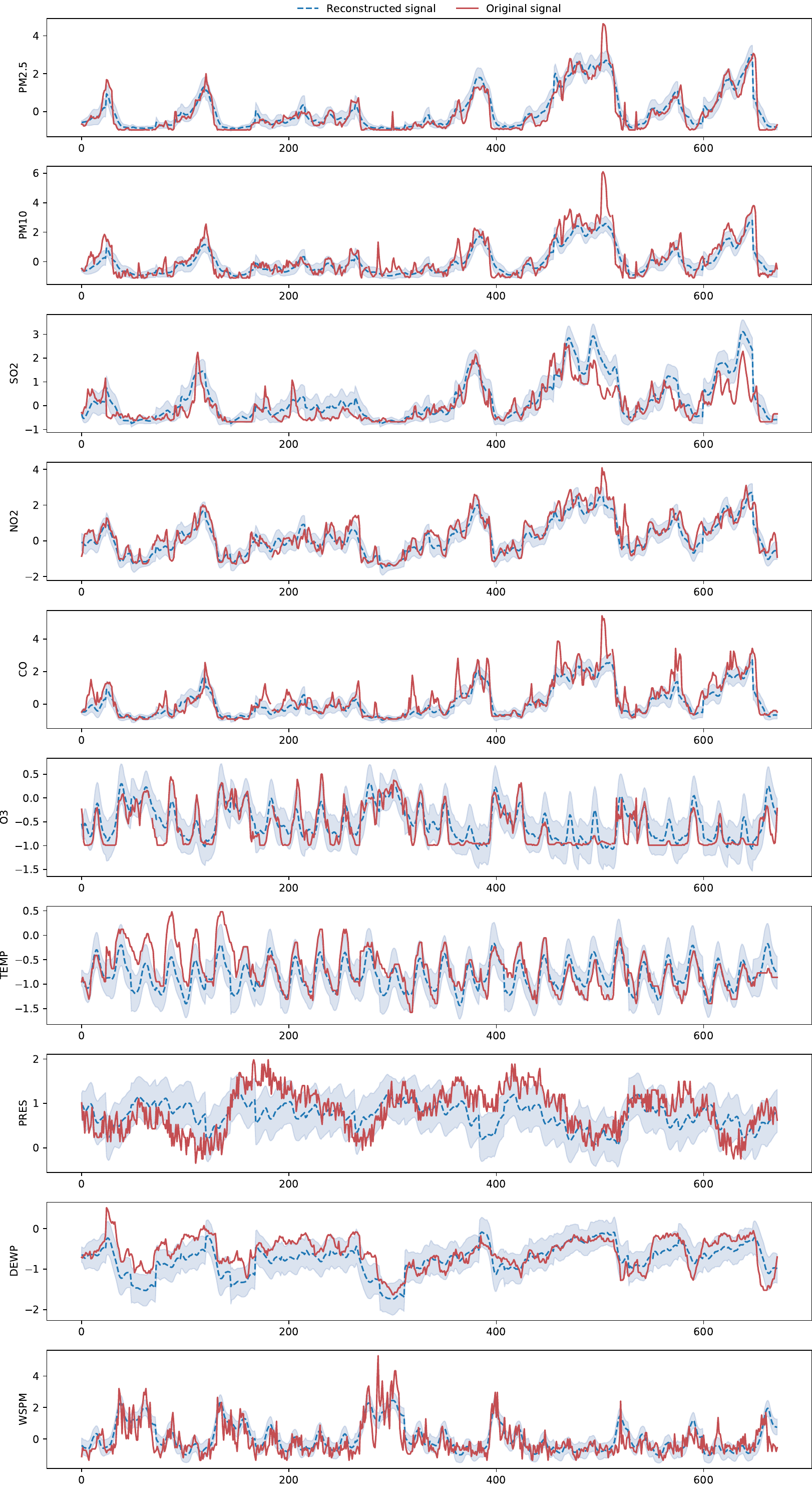}
    \caption{Reconstructed signal of Air Quality by our model}
    \label{fig:recon_ts}
\end{figure}

We present a qualitative analysis of our model reconstruction abilities of time series signals. We use the Air Quality dataset in this analysis. This dataset is comprised of multiple features such as Temperature (TEMP), Carbon Monoxide (CO) and other physical environmental features. In Fig.~\ref{fig:recon_ts}, each plot represents a different feature. The X axis of each plot is the measurement of a specific measure in a specific day. The Y axis are the measurement values. Notably, we observe from this experiment, that our model successfully captures the semantics of each of the time series features. Which in turn, explicitly implies that the latent features of our model encapsulate valuable information about the observed data.

\end{document}